\documentclass[11pt]{article}
\usepackage[margin=1in]{geometry}
\usepackage{amsmath}
\usepackage{amssymb}
\usepackage{bm}
\usepackage{graphicx}
\usepackage{booktabs}
\usepackage{longtable}
\usepackage{array}
\usepackage{pdflscape}
\usepackage{caption}
\usepackage{float}
\usepackage{enumitem}
\usepackage{xcolor}
\usepackage{hyperref}
\usepackage{url}
\usepackage[nopatch=footnote]{microtype}
\usepackage{authblk}
\hypersetup{colorlinks=true, linkcolor=blue, citecolor=blue, urlcolor=blue}
\renewcommand{\arraystretch}{1.12}

\title{Open datasets and machine learning for two-phase heat transfer: a review following a spatial-temporal taxonomy}
\author{\parbox{0.96\textwidth}{\centering
Christy Dunlap\textsuperscript{1}, Ridwan Olabiyi\textsuperscript{2}, Firas Al-Hindawi\textsuperscript{3,4,*}, Hari Pandey\textsuperscript{1}, Stephen Pierson\textsuperscript{1}, Daniel Curl\textsuperscript{1}, Braden Stevens\textsuperscript{1}, Mohammad Ishraq Hossain\textsuperscript{1}, Annapurna Parjuli\textsuperscript{1}, Chinmaya Joshi\textsuperscript{1}, Ashif Iquebal\textsuperscript{2,*}, and Han Hu\textsuperscript{1,*}\par\vspace{0.7ex}
\small
\textsuperscript{1}Department of Mechanical Engineering, University of Arkansas, Fayetteville, AR 72701, USA\\
\textsuperscript{2}School of Computing and Augmented Intelligence, Arizona State University, Tempe, AZ 85281, USA\\
\textsuperscript{3}Industrial \& Systems Engineering Department, King Fahd University of Petroleum \& Minerals, Dhahran, 31261, Saudi Arabia\\
\textsuperscript{4}Interdisciplinary Research Center for Smart Mobility \& Logistics, King Fahd University of Petroleum and Minerals, Dhahran, 31261, Saudi Arabia\par\vspace{0.35ex}
\textsuperscript{*}Corresponding authors: firas.hindawi@kfupm.edu.sa; aiquebal@asu.edu; hanhu@uark.edu
}}
\date{}

\begin{document}
\maketitle

\begin{abstract}
Two-phase heat transfer underpins boiling, condensation, immersion cooling, flow boiling, energy conversion, and electronics thermal management, but its coupled interfacial physics make data reuse and model comparison difficult. This narrative review synthesizes open datasets, machine-learning methods, and reusable software for two-phase heat-transfer research, with emphasis on boiling, multimodal sensing, and thermal-management datasets. We organize the review around a spatial-plus-temporal dimensionality taxonomy, denoted S+TD, that classifies data objects by the dimensionality of the measured, simulated, or derived fields, including 0+0D point values, 0+1D time series, 1+1D profiles, 2+0D images, 2+1D videos, 3+0D/3+1D fields, and mixed multimodal records. The taxonomy is used to connect dataset types to AI tasks such as tabular regression, acoustic sequence learning, image segmentation, video analysis, inverse heat-flux reconstruction, surrogate modeling, and multimodal fusion. The review also develops a roadmap for physics-aware open data, including metadata definitions, evidence and reuse-maturity labels, benchmark splits, decoders, baseline models, and community databanks. NED3 resources are discussed as one implementation case within a broader open-data ecosystem rather than as a complete solution. The main conclusion is that progress in two-phase AI now depends as much on findable, decodable, benchmarkable, and physically interpretable data infrastructure as on model architecture.
\end{abstract}
\noindent\textbf{Keywords:} open-source datasets; multimodal sensing; sequence regression; multiphase transport; boiling heat transfer; thermal management; spatial-temporal data; machine learning; image segmentation

\section{Introduction}

\begin{figure}[H]
\centering
\includegraphics[width=0.95\linewidth]{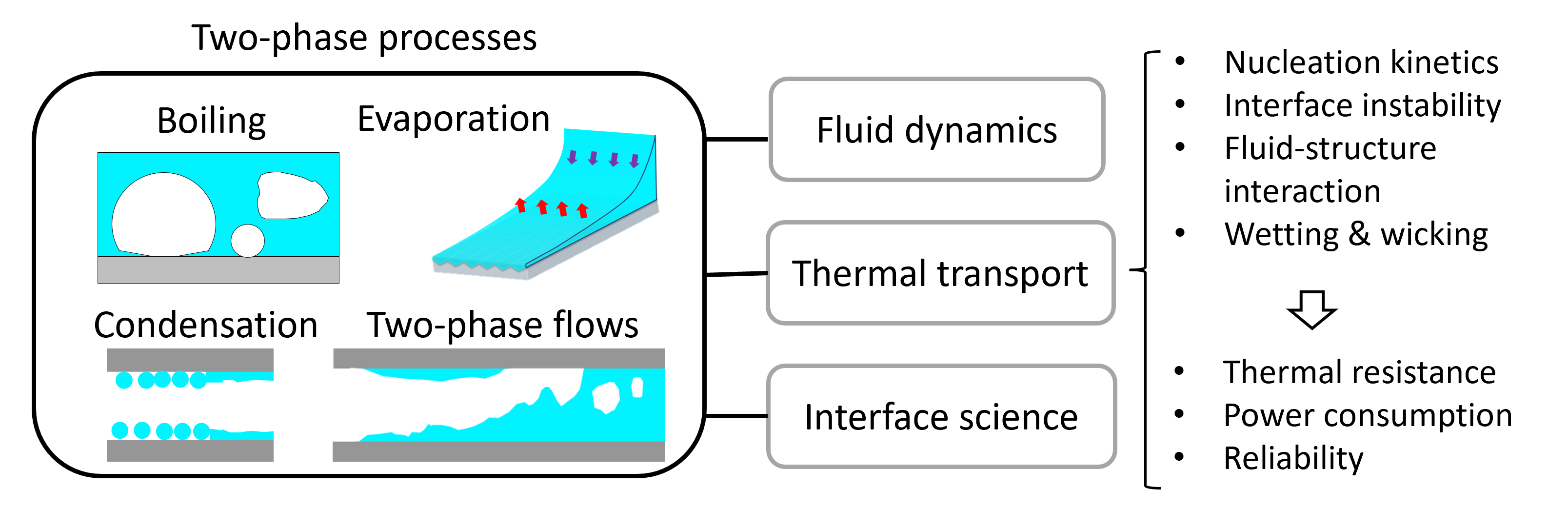}
\caption{Two-phase heat transfer and associated physical processes.}
\label{fig:transport_processes}
\end{figure}

Liquid-vapor two-phase transport is central to energy conversion, electronics cooling, immersion cooling, flow boiling, condensation, and thermal management because phase change can move large heat loads through latent heat and interfacial transport [1-7]. The underlying physics are inherently coupled. As illustrated in Figure~\ref{fig:transport_processes}, boiling and condensation involve nucleation, bubble or droplet growth, coalescence, departure, rewetting, liquid supply, vapor removal, dryout, and wall conduction [1-8]. These mechanisms interact strongly, and performance metrics such as heat-transfer coefficient, critical heat flux, flow regime, and thermal resistance depend on more than one local event or operating variable.

\begin{figure}[H]
\centering
\includegraphics[width=0.95\linewidth]{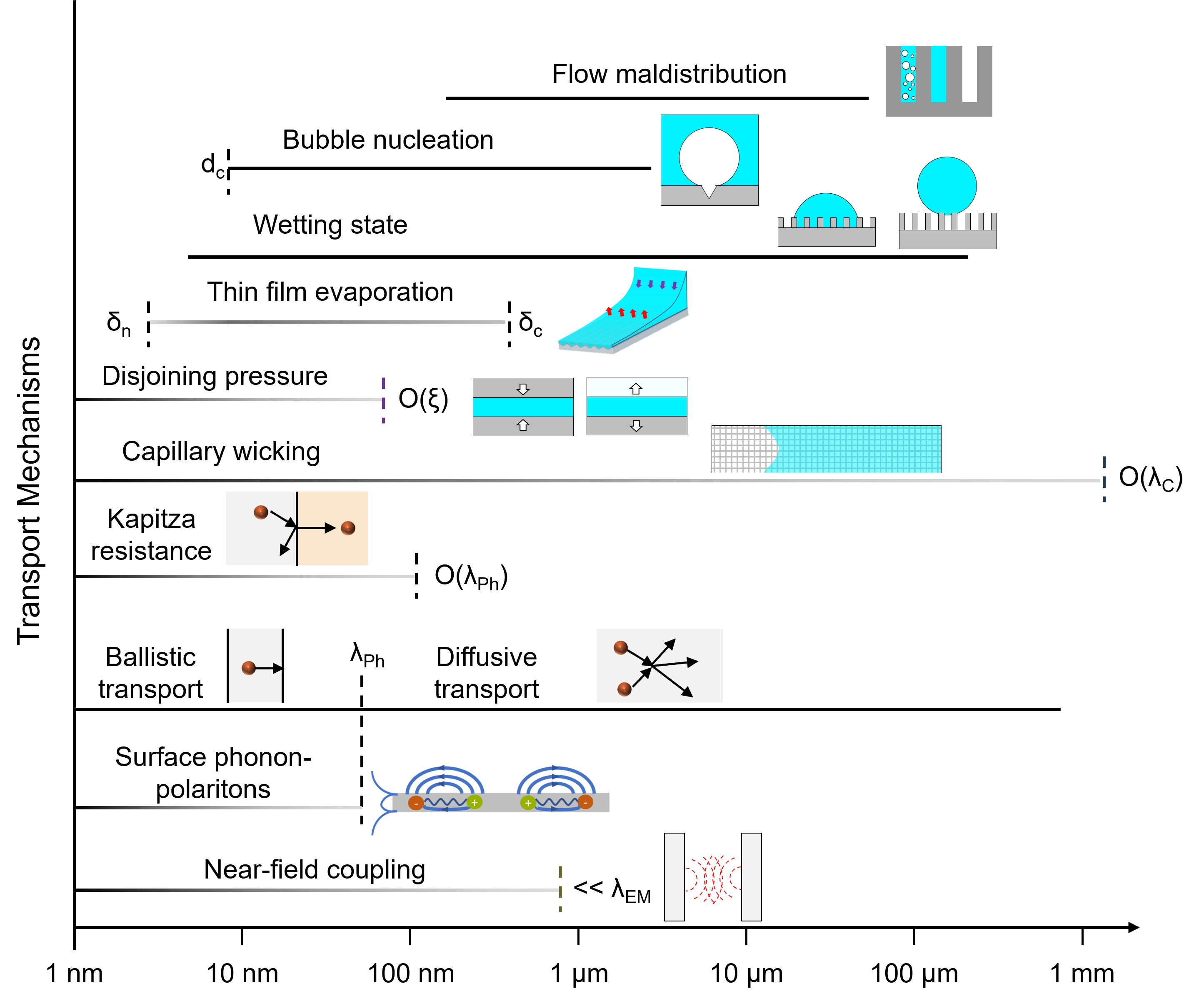}
\caption{Multiscale nature of thermal transport processes spanning molecular, interfacial, component, and system scales.}
\label{fig:multiscale_transport}
\end{figure}

The multiscale nature of heat transfer adds a second difficulty for systematic study and mechanistic modeling. Figure~\ref{fig:multiscale_transport} illustrates molecular transport, interfacial motion, mesoscale bubble or droplet dynamics, component-level flow and conduction paths, and device-level operating conditions. Closure models developed at one scale often become boundary conditions or constitutive inputs at another scale, which makes purely physics-based modeling difficult when surface morphology, transient interfaces, contact-line dynamics, turbulence, evaporation, condensation, and conjugate heat transfer are all active [1-8].

This coupled multiphysics and multiscale structure motivates the data problem addressed in this review. Individual modalities of data, such as temperature, pressure drop, or time-averaged heat flux, can summarize one projection of a two-phase experiment, but they cannot fully describe where bubbles nucleate, how liquid rewets a surface, when vapor blankets form, how acoustic signatures evolve, or how temperature fields reorganize before boiling crisis [8-19]. Data-driven modeling is therefore most useful when it is built on multimodal datasets that probe different projections of the same transport process across scales. Recent studies have been increasingly combining high-speed optical imaging, infrared thermometry, acoustic and acoustic-emission sensing, pressure and temperature histories, surface characterization, and simulation-derived fields for probing two-phase transport [9-28].

Artificial intelligence and machine learning are useful in this setting when they convert heterogeneous measurements into physically meaningful descriptors, predictions, or inverse estimates. Computer vision has been used to extract bubble statistics, interface profiles, void fraction, dry-area information, and flow-regime features from optical and infrared images [9-13,20-23,29-34]. Acoustic and acoustic-emission learning has been used for boiling-regime identification, boiling-crisis prediction, and nonintrusive heat-flux estimation when optical access is limited [14-19]. Scientific machine learning and simulation datasets support optical flow, neural operators, spatiotemporal forecasting, inverse reconstruction, and surrogate modeling for boiling and liquid-cooling systems [35-51]. These examples show that the central question is no longer whether AI can be applied to two-phase heat transfer, but whether the underlying data are discoverable, decodable, comparable, and physically interpretable.

One major practical barrier is that two-phase experiments rarely produce a single clean table. A single test may generate high-speed videos, thermocouple histories, heat-flux estimates, microphone recordings, hydrophone waveforms, acoustic-emission hits, infrared camera files, pressure histories, CAD or geometry files, surface images, and metadata describing fluid, pressure, heat flux, surface condition, sampling rate, sensor location, calibration, and synchronization [24-26,50,52-54]. These signals often use different native formats, clocks, data rates, and processing levels. Some are stored in vendor-specific formats that require proprietary software before analysis can begin. Without physics metadata and open processing logic, a dataset can be public but still difficult to reproduce, benchmark, or combine across laboratories [55-71].

Open datasets and open-source packages provide a means to address this barrier. FAIR data principles, data-citation practices, dataset datasheets, model cards, self-describing scientific formats, and reproducibility guidelines provide general infrastructure for reusable data and models [55-71]. Two-phase heat-transfer data also require domain-specific context, including geometry, fluid properties, surface condition, operating regime, uncertainty, synchronization, and dimensionless groups. A software package without openly available benchmark data may be difficult to validate, while a dataset without decoders or baseline workflows may be technically public but practically inaccessible. The NED3 Laboratory is discussed in this review as one implementation case because its software page describes packages for data acquisition, curation, analysis, and device or system design, while its dataset page summarizes open datasets hosted through Dryad, Dataverse, OSF, and MultiphaseHub [72-73]. The broader point is not that one laboratory provides a complete solution, but that paired release of data and tools is becoming a practical requirement for reproducible two-phase AI.

The literature and repository corpus for this review was assembled during manuscript preparation and updated during revision through July 2026. The search used Google Scholar, publisher pages, arXiv, Crossref/DOI records, GitHub, Zenodo, Hugging Face, Dryad, Harvard Dataverse, Mendeley Data, laboratory dataset pages, and community hubs. Representative search strings combined two-phase or phase-change terms with data and AI terms, including ``boiling machine learning dataset,'' ``two-phase heat transfer open data,'' ``critical heat flux artificial intelligence,'' ``infrared boiling machine learning,'' ``acoustic boiling crisis,'' ``bubble segmentation dataset,'' ``flow boiling machine learning,'' ``multiphase benchmark,'' ``BubbleML,'' ``Bubbleformer,'' ``inverse heat conduction machine learning,'' and ``governing equation discovery thermal fluid.'' Searches were supplemented by backward and forward citation tracing around major clusters in optical and IR boiling diagnostics, acoustic boiling-crisis studies, tabular HTC/CHF modeling, bubble-image analysis, simulation benchmarks, inverse modeling, and scientific machine learning.

Studies and repositories were included when they satisfied at least one of the following criteria: they reported public or reusable two-phase heat-transfer data; released code, decoders, models, or benchmark tasks; used a dataset structure that illustrates an S+TD class; provided representative AI/ML methodology for two-phase or closely related thermal-fluid data; or supplied general open-science infrastructure relevant to dataset curation and benchmarking. Resources were excluded or treated only as contextual examples when they lacked technical documentation, were unrelated to transport physics despite superficially relevant names, could not be connected to a two-phase or thermal-fluid task, or provided no usable information about data modality, labels, model task, or access. Repositories were judged by public accessibility, documentation, persistent archive or versioning, raw-data availability, decoder or loader support, baseline-task definition, and relevance to thermal-fluid physics. The review is therefore narrative and taxonomy-driven rather than PRISMA-style systematic; its aim is to synthesize dataset and model infrastructure across S+TD classes, not to count every paper in each subtopic.

This review is structured around the S+TD taxonomy. First, it introduces S+TD as a compact framework for classifying thermal-fluid datasets by spatial and temporal dimensionality. Second, it organizes available open datasets and repositories, including NED3 resources, by their S+TD classes. Third, it reviews AI/ML methods according to the same S+TD classes, so that new datasets (not included in the present review) can also be aligned to relevant tasks such as regression, segmentation, classification, surrogate modeling, and multimodal fusion. Fourth, it discusses representative open-source tools within the dataset and method review because software is the layer that makes heterogeneous data decodable and reusable. Finally, it treats benchmark datasets, metadata standards, databanks, and one-stop tool libraries as future community infrastructure that should grow from this taxonomy rather than compete with it as a separate organizing theme.

\section{Spatial-plus-temporal dimensionality of thermal-fluid data}

Thermal-fluid datasets can be organized by the dimensionality of the measured, simulated, or derived quantity. We denote this structure as S+TD, where S is the number of spatial dimensions and T is the number of temporal dimensions. The notation is intentionally simple: it describes the shape of a data object from the perspective of analysis and machine learning, not the full dimensionality of the underlying physics. For example, a thermocouple measurement in a boiling experiment is a 0+1D signal even though the temperature field around the heater is spatially distributed. A high-speed video of the same experiment is 2+1D because each frame resolves two spatial dimensions and the sequence evolves in time. A steady CFD temperature field may be 3+0D, while a transient CFD calculation may be 3+1D.

This notation helps clarify why multimodal thermal-fluid learning problems are difficult. Different modalities occupy different S+TD classes, have different sampling rates, and require different preprocessing methods. A hydrophone waveform may be sampled at hundreds of kilohertz or higher, while a heat-flux label may be computed at a much lower rate. A high-speed camera may produce 2+1D optical data at hundreds or thousands of frames per second, while infrared thermography may produce lower-rate but spatially resolved temperature fields. A CFD dataset may provide dense 3+0D or 3+1D fields, but only for a finite set of geometries and boundary conditions. Combining these modalities requires choices about synchronization, resampling, feature extraction, uncertainty, and train-test splitting.

The S+TD framework also provides a neutral vocabulary for connecting datasets to AI/ML tools. Classical regression models are often sufficient for 0+0D metadata or small tabular datasets. Sequence models and spectral features are natural for 0+1D time series. Convolutional neural networks, segmentation models, and vision transformers can be applied to 2+0D images. Recurrent, temporal-convolutional, transformer, or frame-sequence models can be applied to 2+1D videos. Reduced-order models, autoencoders, graph neural networks, neural operators, and physics-informed surrogates may be appropriate for 3+0D or 3+1D fields, depending on the geometry and governing physics. The notation therefore supports both data discovery and model selection.

Several edge cases are worth making explicit. S+TD classifies the data object used by the learning algorithm, not necessarily the physical process that generated it. A spectrogram computed from a hydrophone signal may be represented as a 2+0D image for a CNN, even though it originates from a 0+1D waveform. Event-camera data may be treated as asynchronous 2+1D streams, voxelized event volumes, or frame-like accumulations depending on preprocessing. A line of thermocouples sampled over time is a sparse 1+1D sensor record, while PIV, IR thermography, and optical-flow fields are 2+0D or 2+1D vector/scalar fields depending on whether time is retained. Reduced-order coefficients from POD or autoencoders are not assigned the dimensionality of the original CFD field; they are compact 0+1D or 0+0D representations unless the decoder and spatial basis are included. Point clouds, graphs, and unstructured meshes should be described by both their physical field dimensionality and their computational representation, because the same 3D field may be learned by a neural operator, graph neural network, or reduced-order surrogate. These rules prevent the taxonomy from becoming a rigid file-format label.

\subsection{Physical quantities and metadata definitions}

Because the proposed metadata fields include thermal-hydraulic quantities and dimensionless groups, Table~\ref{tab:physical_metadata_definitions} summarizes common definitions used in boiling and two-phase heat-transfer datasets. The table is not intended to impose a single nomenclature on all subfields. Instead, it identifies quantities that should be explicitly defined in any reusable dataset because definitions can vary with saturation, subcooling, spatial averaging, and flow configuration.

\begingroup
\small
\setlength{\tabcolsep}{4pt}
\renewcommand{\arraystretch}{1.15}
\begin{longtable}{>{\raggedright\arraybackslash}p{0.18\textwidth} >{\raggedright\arraybackslash}p{0.27\textwidth} >{\raggedright\arraybackslash}p{0.47\textwidth}}
\caption{Representative physical quantities and dimensionless metadata for two-phase heat-transfer datasets.}\label{tab:physical_metadata_definitions}\\
\toprule
Quantity & Common definition or meaning & Metadata note \\
\midrule
\endfirsthead
\caption[]{Representative physical quantities and dimensionless metadata for two-phase heat-transfer datasets. (continued)}\\
\toprule
Quantity & Common definition or meaning & Metadata note \\
\midrule
\endhead
\midrule\multicolumn{3}{r}{Continued on next page}\\
\endfoot
\bottomrule
\endlastfoot
Heat flux, $q''$ & Heat-transfer rate per heated area, often based on electrical input after heat-loss correction. & State heated area, heat-loss correction, averaging window, and whether the value is imposed, measured, or inferred. \\
Wall superheat, $\Delta T_w$ & For saturated boiling, commonly $T_w-T_{sat}$; for subcooled boiling, the reference temperature may be $T_{sat}$ or bulk-liquid temperature depending on convention. & Report the reference temperature, wall-temperature location, and spatial averaging method. \\
Heat-transfer coefficient, HTC or $h$ & For saturated boiling, commonly $h=q''/(T_w-T_{sat})$; modified definitions may be used for subcooled, spatially nonuniform, or transient data. & Define wall and reference temperatures, averaging area, and transient window. \\
Critical heat flux, CHF & Maximum sustainable heat flux before boiling crisis or dryout under a specified protocol. & Report detection criterion, heat-flux ramp, dwell time, surface history, pressure, and uncertainty. \\
CHF margin & Distance from boiling crisis, often expressed as $(q''_{CHF}-q'')/q''_{CHF}$ or $q''/q''_{CHF}$. & State whether the margin is normalized as remaining margin or fraction of CHF. \\
Pressure drop, $\Delta p$ & Pressure difference between two taps or locations in a channel or loop. & Report tap locations, elevation correction, single- or two-phase contribution, and sensor uncertainty. \\
Vapor quality, $x$ & Mass fraction of vapor in a two-phase mixture under equilibrium or model assumptions. & State whether quality is thermodynamic equilibrium quality, local quality, or inferred from an energy balance. \\
Void fraction, $\alpha$ & Fraction of volume, area, or line of sight occupied by vapor, depending on diagnostic method. & Specify volume-, area-, chord-, image-, or time-averaged definition and thresholding method. \\
Regime label & Discrete description such as bubbly, slug, churn, annular, nucleate boiling, transition boiling, film boiling, dryout, or rewetting. & Provide label ontology, annotator or algorithm, transition rules, and ambiguity handling. \\
Reynolds number, $Re$ & Ratio of inertial to viscous effects, typically $\rho U L/\mu$ with geometry-specific choices of $U$ and $L$. & State phase, mixture model, velocity, length scale, and property-evaluation temperature. \\
Weber number, $We$ & Ratio of inertial to surface-tension effects, commonly $\rho U^2 L/\sigma$. & State velocity and length scale; use phase-specific properties when needed. \\
Jakob number, $Ja$ & Ratio of sensible heat to latent heat, commonly $c_p\Delta T/h_{fg}$. & State whether $\Delta T$ is wall superheat, subcooling, or another temperature difference. \\
Boiling number, $Bo$ & Ratio of heat input to convective latent transport, commonly $q''/(G h_{fg})$ in flow boiling. & Report mass flux $G$, latent heat source, and area basis. \\
Bond number, $Bd$ or $Bo_d$ & Ratio of buoyancy to surface-tension effects, commonly $\Delta\rho g L^2/\sigma$. & Avoid ambiguity with Boiling number by defining the symbol and length scale. \\
Capillary number, $Ca$ & Ratio of viscous to surface-tension effects, commonly $\mu U/\sigma$. & State phase viscosity and characteristic velocity. \\
Confinement number, $Co$ & Ratio comparing capillary length to channel length scale, often involving $[\sigma/(g\Delta\rho)]^{1/2}/D_h$. & Define hydraulic diameter or channel dimension and gravity orientation. \\
Reduced pressure, $p_r$ & Pressure normalized by critical pressure, $p/p_c$. & State fluid property source and operating pressure. \\
CHF-normalized heat flux & Heat flux normalized by CHF, Zuber limit, kinetic limit, or another reference. & State the reference model or measured CHF and its validity range. \\
\end{longtable}
\endgroup

\subsection{S+TD taxonomy}

\begin{figure}[H]
\centering
\includegraphics[width=0.95\linewidth]{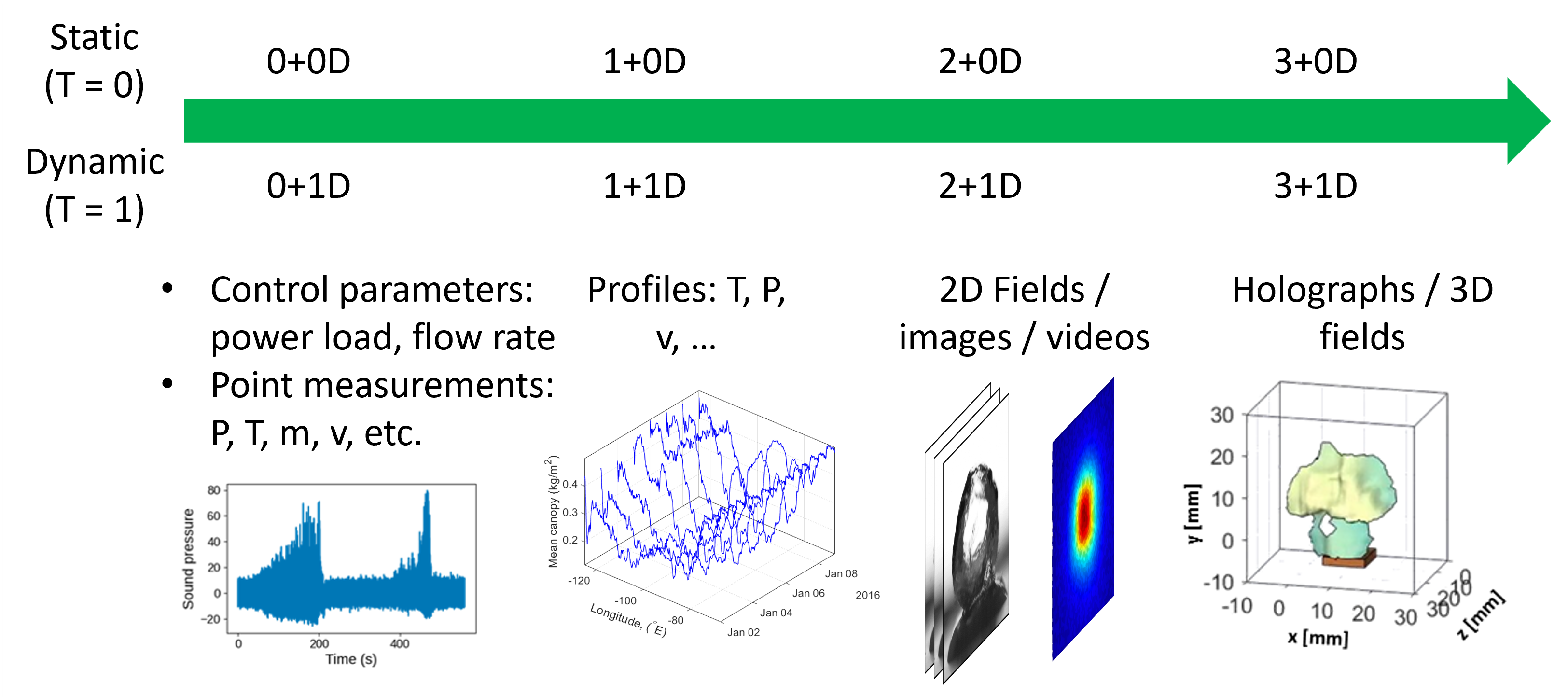}
\caption{S+TD taxonomy for thermal-fluid datasets. Each spatial class includes static and time-dependent examples: 0+0D point measurements and 0+1D sensor histories; 1+0D profiles and 1+1D evolving profiles; 2+0D images or fields and 2+1D videos or field sequences; and 3+0D steady volumes or fields and 3+1D transient volumes. Mixed S+TD records combine synchronized modalities such as optical video, acoustic waveforms, IR fields, heat-flux labels, and physics metadata.}
\label{fig:std_taxonomy}
\end{figure}

Figure~\ref{fig:std_taxonomy} illustrates the S+TD taxonomy with representative signals from boiling studies, ranging from point measurements to time series, images, videos, and three-dimensional fields for both static and dynamic measurements. Table~\ref{tab:1} summarizes representative S+TD classes relevant to open thermal-fluid datasets, related machine learning tasks, and additional information required for effective reuse.

\begingroup
\footnotesize
\setlength{\tabcolsep}{3pt}
\renewcommand{\arraystretch}{1.18}
\begin{longtable}{>{\raggedright\arraybackslash}p{0.10\textwidth} >{\raggedright\arraybackslash}p{0.15\textwidth} >{\raggedright\arraybackslash}p{0.27\textwidth} >{\raggedright\arraybackslash}p{0.24\textwidth} >{\raggedright\arraybackslash}p{0.18\textwidth}}
\caption{S+TD taxonomy for thermal-fluid datasets.}\label{tab:1}\\
\toprule
Class & Data object & Thermal-fluid examples & Common AI/ML tasks & Notes for reuse \\
\midrule
\endfirsthead
\caption[]{S+TD taxonomy for thermal-fluid datasets. (continued)}\\
\toprule
Class & Data object & Thermal-fluid examples & Common AI/ML tasks & Notes for reuse \\
\midrule
\endhead
\midrule\multicolumn{5}{r}{Continued on next page}\\
\endfoot
\bottomrule
\endlastfoot
0+0D & Static scalar or categorical value & Fluid type, surface roughness, steady mean heat flux, nominal power, geometry label & Tabular regression, classification, metadata filtering & Useful as metadata or labels; rarely sufficient alone for transient multiphase physics. \\
0+1D & Point time series & Thermocouple history, pressure signal, hydrophone waveform, AE hit sequence, heater power & Sequence regression, event detection, spectral analysis, anomaly detection & Requires sampling rate, filtering, sensor location, calibration, and synchronization information. \\
1+0D & Static profile & Line-scan temperature profile, channel centerline velocity profile, radial heat-flux profile & Profile regression, interpolation, reduced-order descriptors & Requires coordinate definition, spatial resolution, and boundary-condition context. \\
1+1D & Time-evolving profile & Temperature profile over time, line-scan vapor interface, channel pressure distribution over time & Sequence-to-sequence prediction, temporal profile reconstruction & Useful bridge between point signals and full fields. \\
2+0D & Static image or 2D field & Boiling image, IR thermograph, micrograph, surface map, 2D simulation slice & Segmentation, classification, object detection, image regression & Requires field of view, scale, lighting/emissivity, camera model, and annotation definitions. \\
2+1D & Video or time-evolving 2D field & High-speed boiling video, IR video, time-resolved 2D CFD slice & Video regression, event prediction, optical-flow-like analysis, multimodal fusion & Requires frame rate, exposure, synchronization, and train-test split design that avoids leakage across experiments. \\
3+0D & Static volume or 3D field & Holographic reconstruction, 3D CFD temperature/velocity field, 3D surface scan & Surrogate modeling, reduced-order modeling, geometry-aware learning & Requires mesh/voxel representation, coordinate system, boundary conditions, and validation status. \\
3+1D & Time-evolving volume & Transient CFD field, volumetric imaging sequence, 3D digital twin state history & Neural operators, spatiotemporal surrogates, state estimation, control & Data size and generalization risk are high; metadata and compression become critical. \\
Mixed S+TD & Synchronized multimodal record & Optical video plus acoustic waveform plus heat-flux label; IR field plus thermocouples plus power & Sensor fusion, multimodal representation learning, cross-modal prediction & Requires explicit clock alignment, missing-data rules, and modality-specific uncertainty. \\
\end{longtable}
\endgroup

\section{Open datasets organized by the S+TD taxonomy}

Available two-phase AI resources already cover much of this S+TD range, but they are distributed across papers, repositories, and archival platforms. Recent phase-change AI reviews frame the field around data extraction, data-stream analysis, physics-centered machine learning, and sustainable cyberinfrastructure [74]. Infrared thermometry and boiling-crisis studies provide 2+1D thermal-field examples [8-12]. Optical boiling and flow-visualization studies provide 2+0D images, 2+1D videos, segmentation labels, and vision-derived bubble descriptors [13,20-22,29-32,75-81]. Acoustic and acoustic-emission datasets provide 0+1D waveforms, spectrograms, hit sequences, and regime or heat-flux labels [14-18,23,82]. Simulation resources such as BubbleML, BubbleML 2.0, Bubbleformer, NUCLEUS-MoE, HB-ARFM, and bubble-dynamics repositories provide controlled 0+1D, 2+1D, 3+0D, and 3+1D data for surrogate modeling, operator learning, forecasting, unified pool-boiling surrogates, inverse reconstruction, and scientific ML [19,35-43,46-50]. Additional image-analysis, multimodal, and NED3 datasets extend the landscape through interpretability tools, synchronized optical-acoustic-thermal measurements, bubble-image archives, optical-to-infrared condensation mapping, hydrophone/AE heat-flux data, image sequences, and boiling-hysteresis records [24-26,51-54,83-86]. Software-linked resources, including SeqReg, Data Center Cooling Research Tools, BoilingBench-Multimodal, Multiphase Hub, and Bubble Dynamics Research Tools, show that dataset release is increasingly tied to reusable workflows and community curation [87-91].

Table~\ref{tab:2} summarizes representative public datasets and repositories using S+TD classes as the organizing language. This table asks which AI tasks are plausible for a given dimensionality and modality, what metadata are needed, and what benchmark or tool support is already available. To reduce ambiguity, the final column labels both evidence level and reuse maturity. Evidence level distinguishes peer-reviewed datasets or papers, archival records, maintained software, benchmark suites, educational examples, and proposed seed infrastructure. Reuse maturity is described qualitatively as unclear, emerging, partial, or mature according to public access, documentation, versioning, raw-data availability, decoder support, baseline tasks, and independent reproducibility.

\begin{landscape}
\begingroup
\scriptsize
\setlength{\tabcolsep}{3pt}
\renewcommand{\arraystretch}{1.18}
\begin{longtable}{>{\raggedright\arraybackslash}p{0.18\linewidth} >{\raggedright\arraybackslash}p{0.21\linewidth} >{\raggedright\arraybackslash}p{0.12\linewidth} >{\raggedright\arraybackslash}p{0.17\linewidth} >{\raggedright\arraybackslash}p{0.24\linewidth}}
\caption{Representative public datasets and repositories for AI-assisted two-phase analysis.}\label{tab:2}\\
\toprule
Resource & Modality or method & Representative S+TD class & Main use & Evidence level and reuse maturity notes \\
\midrule
\endfirsthead
\caption[]{Representative public datasets and repositories for AI-assisted two-phase analysis. (continued)}\\
\toprule
Resource & Modality or method & Representative S+TD class & Main use & Evidence level and reuse maturity notes \\
\midrule
\endhead
\midrule\multicolumn{5}{r}{Continued on next page}\\
\endfoot
\bottomrule
\endlastfoot
Hobold et al. visualization studies [13,75] & Low-speed direct/indirect boiling visualization, SVM, neural networks, PCA/MLP & 2+0D and 2+1D & Regime classification and heat-flux estimation from images & Peer-reviewed studies; reuse maturity unclear to partial. Useful historical baseline showing that compact image features can support nonintrusive diagnostics. \\
Suh et al. visual boiling framework [76] & High-speed images, CNN, object detection, bubble statistics & 2+0D and 2+1D & Heat-flux prediction from visual bubble features & Peer-reviewed study; reuse maturity partial. Strong example of physically reinforced vision learning; benchmark reuse depends on data and annotation availability. \\
Suh et al. VISION-iT framework [32] & Mask R-CNN, tracking, postprocessed bubble/droplet descriptors & 2+1D & Large-scale extraction of nucleation dynamics and bubble/droplet features & Peer-reviewed framework; reuse maturity partial. Important pipeline model for converting videos into physically interpretable descriptors and labels. \\
Fu et al. Bubble2Heat framework [21] & Segmented high-speed recordings, thermocouple readings, simulation-trained generative AI & Mixed 2+1D and thermal-field inference & Optical-to-thermal field reconstruction & Peer-reviewed cross-modal framework; reuse maturity emerging. Strong example of cross-modal inference; databank reuse requires aligned simulation, optical, and temperature metadata. \\
Optical-to-infrared condensation mapping [85-86] & Paired optical/infrared condensation data, cGAN-based image-to-temperature mapping, vapor-pressure conditioning & 2+1D optical and thermal-field sequences & Image-to-temperature inference and local condensation heat-transfer analysis & Peer-reviewed code/sample-data artifact; reuse maturity partial. Important cross-modal example outside boiling; public code and sample data help, but full benchmark reuse would require complete paired data, calibration, split files, and uncertainty metadata. \\
Chang et al. EventFlow framework [22] & Neuromorphic event data, LSTM and related classifiers & Event-based 2+1D & Real-time flow-regime classification & Peer-reviewed emerging diagnostic; reuse maturity emerging. Useful emerging example of low-latency AI diagnostics for flow boiling. \\
Huang et al. flow-boiling visualization study [77] & Flow visualization and ML-extracted physical features & 2+1D & Void fraction, HTC, CHF model support & Peer-reviewed study; reuse maturity emerging. Shows AI as a feature-extraction engine for theoretical model improvement. \\
Consolidated flow-boiling database models [20,78] & Literature databases, dimensionless groups, ANN/tree-based regressors & 0+0D/tabular & Flow-boiling pressure-drop and HTC prediction & Peer-reviewed tabular databases; reuse maturity partial. Demonstrates why curated tabular databanks remain important beside image/video archives. \\
IR boiling-crisis studies [8-12] & High-speed IR thermometry, dry-area/bubble-footprint analysis, neural networks & 2+1D thermal fields & Distance-to-crisis prediction, dryout detection, boiling-crisis mechanism analysis & Peer-reviewed studies; reuse maturity partial. Needs detailed calibration, heater geometry, inverse methods, and uncertainty metadata. \\
Acoustic boiling-crisis datasets and models [15-17,24,82] & Acoustic emission/audio spectrograms and CNN classification & 0+1D transformed to time-frequency images & Boiling-regime and crisis prediction & Peer-reviewed datasets/models; reuse maturity partial. Strong visual-limited diagnostic path; sensor and noise metadata are essential. \\
Boiling acoustic-emission Data in Brief dataset [15] & AE data for data-driven boiling-regime prediction & 0+1D & Regime classification and acoustic diagnostics & Peer-reviewed archival dataset; reuse maturity mature for acoustic-regime tasks. Useful for acoustic benchmarks and cross-lab comparison. \\
BubbleML [19,35] & Physics-driven boiling simulations and benchmarks & 2+1D, 3+0D/3+1D depending representation & Optical flow, operator learning, spatiotemporal prediction & Peer-reviewed simulation benchmark; reuse maturity mature for documented simulation tasks. Strong simulation benchmark; should be paired with experimental validation tasks. \\
Bubbleformer and BubbleML 2.0 [40-43] & Transformer forecasting model, Flash-X boiling simulations, HDF5 fields, thermophysical parameters & 2+1D and 3+1D simulation-derived fields & Long-range boiling forecasting, cross-fluid generalization, physics-based benchmark metrics & Peer-reviewed/preprint benchmark ecosystem; reuse maturity emerging to partial. Important expansion from prediction toward autonomous forecasting; curation should preserve solver settings, fluid properties, boundary conditions, split files, and physical-fidelity metrics. \\
NUCLEUS-MoE and NUCLEUS package [49-50] & Mixture-of-experts surrogate, neighborhood attention, interface-consistency treatment, expert routing & 2+1D and 3+1D simulation-derived pool-boiling fields & Unified saturated/subcooled pool-boiling surrogate modeling, cross-fluid and few-shot generalization & Peer-reviewed/preprint software artifact; reuse maturity emerging. Relevant to benchmark design because it targets generalization across dielectrics, refrigerants, cryogens, thermodynamic conditions, and immersion-cooling fluids. \\
HB-ARFM inverse boiling reconstruction [46-47] & History-bootstrapped autoregressive flow matching, partial observations from interface geometry/motion, public repository/release path & 2+1D observations to 2+1D/3+1D field reconstruction & Spatiotemporal inverse reconstruction of velocity and temperature fields from imaging-derived observations & Preprint/software artifact; reuse maturity emerging. Highlights an emerging benchmark direction in which visual records become conditioning data for hidden thermofluid-state reconstruction. \\
FD-Bench two-phase boiling evaluation [48] & Modular data-driven fluid-simulation benchmark using BubbleML pool- and flow-boiling cases & 2+1D and 3+1D simulation-derived fields & Cross-architecture comparison for data-driven fluid simulation & Benchmark-suite artifact; reuse maturity emerging to partial. Shows that two-phase boiling data can become part of broader SciML benchmark suites; benchmark papers should preserve dataset provenance, physical metrics, and two-phase-specific failure modes. \\
BubMask [79-80] & Mask R-CNN bubble segmentation & 2+0D and 2+1D & Bubble masks, centroids, areas, axes, orientation & Open-source software; reuse maturity partial. Mature example of open bubble-image segmentation; dependency age and model portability should be checked. \\
BubbleID [29-30,81] & Mask R-CNN/SORT-style segmentation, tracking, classification & 2+0D and 2+1D & Bubble statistics, interface dynamics, departure classification & Peer-reviewed dataset/software co-release; reuse maturity partial to mature. Dataset/software co-release pattern directly aligned with the paired data-and-tool model reviewed here. \\
BubbleNet [36] & Physics-informed neural networks for microbubble flow & Simulation fields & Bubble-flow inference and field prediction & Simulation/modeling repository; reuse maturity emerging. Useful as a physics-informed modeling candidate; not a direct experimental boiling benchmark. \\
FluidNexus [31] & Single-video fluid reconstruction and prediction & 2+1D to 3D reconstruction & 3D fluid reconstruction, video prediction & Broader fluid-AI repository; reuse maturity emerging. Broader fluid AI example; relevant to future multimodal/video-based thermal-fluid models. \\
bubble-dynamics-resnet [37] & ResNet surrogate for Rayleigh-Plesset ODE data & 0+1D simulation & Accelerated bubble-dynamics integration & Educational/scientific-ML repository; reuse maturity emerging. Archived educational/scientific ML example; evidence level should be labeled. \\
BerryWei/Bubble-dynamic [38] & Rayleigh-Plesset and Keller-Miksis simulation & 0+1D simulation & Single-bubble dynamics modeling & Educational/mechanistic code; reuse maturity emerging. Candidate educational tool; needs screening before benchmark use. \\
kozakaron/\\Bubble\_dynamics\_simulation [39] & Bubble dynamics with chemical/thermal modeling & 0+1D simulation & Bubble dynamics and reaction-coupled modeling & Educational/mechanistic code; reuse maturity emerging. Candidate mechanistic code; scope differs from boiling-image datasets. \\
JackEdTaylor/bubblemask [51] & Gaussian "Bubbles" image masks & 2+0D image perturbation & Visual interpretability and masked-image analysis & General image-interpretability code; reuse maturity partial outside two-phase physics. Not a two-phase bubble tool despite the name, but useful for image-interpretability methods. \\
Harvard Dataverse multimodal boiling dataset [83-84] & Optical, hydrophone, AE, thermal measurements & Mixed S+TD & Multimodal boiling and CHF monitoring & Archival multimodal dataset; reuse maturity partial to mature. Key example of synchronized acoustic-optical-thermal release. \\
NED3 Dryad acoustic/image datasets [17-18,24-26,30,52-54,83-84] & Bubble images, hydrophone, AE, high-speed videos, thermal labels & 0+1D, 2+0D, 2+1D, mixed & Heat-flux regression, segmentation, hysteresis, multimodal fusion & Archival datasets and software-linked releases; reuse maturity partial, with benchmark readiness under development. Strong basis for benchmark tasks if common splits and metadata schemas are added. \\
\end{longtable}
\endgroup
\end{landscape}

Table~\ref{tab:2} also exposes several field-level gaps. First, public availability is uneven. Some resources provide raw files and reusable code, while others provide only papers, derived figures, or partially documented repositories. Second, modality-rich datasets are often not benchmark-ready because synchronization, uncertainty, train-test split logic, and baseline models are missing or incomplete. Third, simulation datasets are valuable for controlled learning tasks but should not be treated as substitutes for experimental validation unless uncertainty, boundary conditions, closure assumptions, and transfer tests are documented. Finally, repositories with similar names or superficial relevance can differ sharply in scientific evidence level. A useful databank must therefore distinguish peer-reviewed datasets, archival records, maintained software, educational examples, preliminary code, and proposed benchmark mechanisms.

\subsection{NED3 datasets as an S+TD case study}

The NED3 dataset collection is used here as a compact case study in paired dataset/software release, not as the center of the review. It includes examples from boiling heat transfer, immersion cooling, thermal-resistance measurement, CFD surrogate modeling, device design, liquid cooling, flow boiling, partial discharge, and related thermal-fluid topics. The public dataset page states that non-proprietary datasets are hosted through external platforms including Dryad, Dataverse, OSF, and MultiphaseHub [73]. This distributed strategy is useful because different datasets require different archival, DOI, metadata, and file-size support, but it also shows why a community databank should help users compare resources across laboratories rather than navigate one laboratory page at a time.

The released datasets are intentionally heterogeneous because thermal-fluid AI cannot be reduced to one benchmark type. Image-heavy datasets support computer vision and interface analysis. Acoustic datasets support sequence regression and nonintrusive monitoring. Multimodal datasets support sensor fusion and cross-modal inference. CFD-generated datasets support surrogate modeling and digital twins. Design-file collections support reproducible device fabrication and experimental repeatability. The S+TD framework makes this diversity navigable without forcing incompatible records into a single data mold.

The papers and archived records behind these datasets also show why a single benchmark style would be too narrow. The hydrophone-based boiling-acoustics study framed acoustic sensing as a nonintrusive route to heat-flux quantification during pool boiling [26]. The temporal-spatial boiling framework extended this idea to efficient heat-flux monitoring from image-derived and time-resolved information [52], while Hit2Flux focused on acoustic-emission hit sequences as inputs for heat-flux prediction [53]. The boiling-hysteresis and multimodal-boiling releases emphasize synchronized acoustic, optical, and thermal measurements for transient and steady-state boiling analysis [25,54,83-84]. The cold-plate CFD/emulation work extends the ecosystem beyond boiling experiments by using POD-NN surrogate modeling for electronics-cooling thermal fields [92]. The immersion-cooler Dryad record similarly broadens the dataset portfolio toward two-phase electronics cooling for medium-voltage power modules [93]. 

\begin{figure}[H]
\centering
\includegraphics[width=0.95\linewidth]{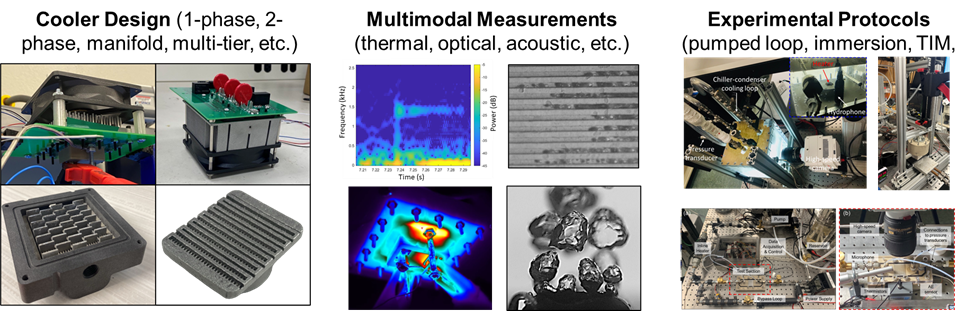}
\caption{Categories of open-source data in the NED3 ecosystem showing that open thermal-fluid resources include test section designs, thermal fluid testbeds, and processed data.}
\label{fig:ned3_data_categories}
\end{figure}

Figure~\ref{fig:ned3_data_categories} shows the breadth of data types that appear in this ecosystem, including raw measurements, design files, instrument outputs, processed data, and experiment documentation. A reusable thermal-fluid dataset often needs this surrounding context, not only the final input-output table used for ML training.

\subsection{Dataset inventory}

Table~\ref{tab:3} summarizes the NED3 datasets listed publicly as of May 21, 2026. Availability status and descriptions are based on the NED3 dataset page [73]. The S+TD assignments are author interpretations based on the reported modalities, rather than official categories assigned by the dataset page.

\begin{landscape}
\begingroup
\scriptsize
\setlength{\tabcolsep}{3pt}
\renewcommand{\arraystretch}{1.18}
\begin{longtable}{>{\raggedright\arraybackslash}p{0.06\linewidth} >{\raggedright\arraybackslash}p{0.16\linewidth} >{\raggedright\arraybackslash}p{0.27\linewidth} >{\raggedright\arraybackslash}p{0.13\linewidth} >{\raggedright\arraybackslash}p{0.18\linewidth} >{\raggedright\arraybackslash}p{0.12\linewidth}}
\caption{NED3 open datasets and representative S+TD classes.}\label{tab:3}\\
\toprule
ID & Dataset & Reported modalities or contents & Representative S+TD classes & Example research uses & Reuse maturity \\
\midrule
\endfirsthead
\caption[]{NED3 open datasets and representative S+TD classes. (continued)}\\
\toprule
ID & Dataset & Reported modalities or contents & Representative S+TD classes & Example research uses & Reuse maturity \\
\midrule
\endhead
\midrule\multicolumn{6}{r}{Continued on next page}\\
\endfoot
\bottomrule
\endlastfoot
ned3-001 & BubbleID & Raw boiling images, annotated images, saved model weights & 2+0D, labels, model artifacts & Bubble segmentation, interface detection, bubble statistics, computer-vision benchmarking.  & Partial to mature: archived data/software/model artifacts. \\
ned3-002 & Boiling Hysteresis & Temperature, hydrophone, microphone, AE hits, optical images during transient pool boiling & 0+1D, 2+0D, mixed S+TD & Hysteresis analysis, multimodal event detection, cross-modal boiling diagnostics.  & Partial: multimodal archive; benchmark splits still needed. \\
ned3-003 & Immersion Cooling Thermograph & Temperature histories and steady-state temperature profiles with thermocouples and FLIR IR imaging & 0+1D, 1+0D, 2+0D & Two-phase immersion cooling, electronics thermal management, IR/thermocouple comparison.  & Partial: public record; benchmark task not standardized. \\
ned3-004 & Boiling Acoustics - Hydrophone & Synchronized hydrophone acoustic and temperature/heat-flux data for transient pool boiling & 0+1D & SeqReg heat-flux prediction, spectral features, nonintrusive boiling diagnostics.  & Partial to mature: archival acoustic dataset; decoder/baseline workflows available through related tools. \\
ned3-005 & Boiling Acoustics - AE Sensor & AE hit data and waveforms, hydrophone data, temperature/heat-flux data for copper microchannels & 0+1D, mixed S+TD & Hit2Flux-style prediction, AE feature learning, transient boiling monitoring.  & Partial: archival AE dataset; run-level benchmark definitions needed. \\
ned3-006 & Boiling Image Sequences & High-speed videos, hydrophone data, temperature/heat-flux data from copper microchannels & 2+1D, 0+1D, mixed S+TD & Image-sequence regression, video-based heat-flux prediction, multimodal fusion.  & Partial: video-acoustic archive; benchmark splits under development. \\
ned3-007 & Multimodal Boiling Data & High-speed videos, hydrophone data, temperature profiles from steady and transient pool boiling on multiple surfaces & 2+1D, 0+1D, 1+0D, mixed S+TD & Benchmarking across surfaces, regime identification, generalization tests.  & Partial: multimodal archive; community benchmark expansion needed. \\
ned3-008 & Modified ASTM D5470 Test Data & Temperature histories from six thermocouples for thermal resistance and conductivity analysis & 0+1D & Uncertainty quantification, thermal-resistance data reduction, educational benchmarks.  & Emerging to partial: useful educational/reduction dataset. \\
ned3-009 & Cold Plate CFD and AI Emulation Data & 2000 steady-state CFD simulations of a dual-chip pin-fin cold plate, POD-NN code, post-analysis code & 3+0D, tabular design variables & CFD surrogate modeling, reduced-order modeling, digital twins for electronics cooling.  & Partial: CFD/surrogate dataset with code; external validation tasks needed. \\
ned3-010 & Design Files for Twisted Pair Preparation Device & Design files and testing demonstrations for hand-operated and motorized devices & 0+0D metadata, CAD/design artifacts & Reproducible test-device fabrication, partial-discharge sample preparation.  & Partial: design-file release rather than ML benchmark. \\
ned3-011 & Hybrid PA12-CF/AlSi10Mg Liquid Cooler for 10 kV Power Module & Thermal-hydraulic data and IR images of multi-tier liquid coolers & 0+1D, 2+0D & Power-module cooling, liquid-cooler performance analysis, IR validation.  & Partial: thermal-hydraulic/IR record; benchmark task not standardized. \\
ned3-012 & Flow Boiling Acoustic Emissions & Acoustic-emission sensing data of single-phase and two-phase flows in a straight microchannel heat sink & 0+1D & Flow-regime diagnostics, AE event classification, two-phase monitoring.  & Emerging to partial: curated AE flow dataset. \\
ned3-013 & Subatmospheric Pressure Boiling & Temperature, pressure, and high-speed videos of pool boiling at 10-100 kPa; listed as being curated & 0+1D, 2+1D & Pressure-dependent boiling dynamics, regime transition analysis.  & Emerging: listed as being curated. \\
ned3-014 & Acoustic Detection of Partial Discharge & AE data of partial discharge in twisted pairs, power modules, and statorettes; listed as being curated & 0+1D & Nonintrusive electrical insulation diagnostics, AE event detection.  & Emerging: listed as being curated. \\
ned3-015 & 3D-Printed Titanium-Encapsulated Pyrolytic Graphite Fin & Thermal test data for radiator fins with and without pyrolytic graphite; listed as being curated & 0+1D, possible 2+0D & Thermal-spreader and radiator-fin evaluation, material-system comparison.  & Emerging: listed as being curated. \\
ned3-016 & Subcooled Pool Boiling & High-speed videos, hydrophone, and thermal data for subcooled pool boiling; listed as being curated & 2+1D, 0+1D, mixed S+TD & Subcooling effects, video-acoustic fusion, boiling-regime learning.  & Emerging: listed as being curated. \\
\end{longtable}
\endgroup
\end{landscape}

\subsection{Dataset metadata needed for reusable AI benchmarks}

The scientific value of an open thermal-fluid dataset depends on metadata as much as file availability. For experimental datasets, users need geometry, fluid, surface preparation, operating pressure, heat-input protocol, calibration information, sensor location, sampling rate, clock synchronization, data-reduction equations, filtering choices, uncertainty estimates, and repeated-test information. For optical and infrared data, users also need field of view, pixel scale, lighting or emissivity assumptions, exposure, frame rate, and preprocessing history. For acoustic data, users need sensor model, coupling, location, amplifier settings, thresholding rules, waveform window length, frequency response, and noise characterization. For CFD datasets, users need geometry, mesh, turbulence or multiphase model, boundary conditions, convergence criteria, material properties, solver version, and validation or sanity-check data. These fields are the engineering context that determines whether a model is learning transport behavior or only facility-specific artifacts.

These metadata requirements are not administrative details; they determine whether a model evaluation is physically meaningful. A random train-test split across frames from the same video may overestimate generalization because adjacent frames share nearly identical conditions. A split across operating cases, surfaces, fluids, or geometries is more demanding and more useful for engineering deployment. Similarly, a model that performs well on synchronized data from one experiment may fail if sensor timing, sampling rate, or filtering differs. Open datasets therefore need benchmark splits and clear warnings about leakage modes.

\section{AI/ML methods organized by S+TD class}

The S+TD taxonomy also provides a practical entry point for selecting machine-learning methods. A user with a scalar boiling database faces a different problem from a user with hydrophone waveforms, high-speed videos, infrared fields, or CFD volumes. Organizing the literature by data dimensionality offers a direct path from dataset type to relevant model families, benchmark tasks, and failure modes.

\subsection{ML methods for 0+0D tabular data}

The 0+0D class includes scalar operating conditions, material descriptors, dimensionless groups, and performance labels. This is the natural format for heat-transfer-coefficient, pressure-drop, void-fraction, flow-regime, and CHF databases. Classical boiling and flow-boiling physics established that these quantities are regime-dependent functions of fluid properties, geometry, pressure, mass flux, heat flux, vapor quality, surface condition, and gravity [1-7,94-108]. This history matters for ML because it defines the physical variables that should appear as features, metadata, or baseline correlations.

Tabular two-phase ML has been used to build data-driven correlations and predictive models using random forests, gradient boosting, Gaussian processes, support-vector machines, neural networks, and related regressors [109-117]. The strongest studies are not merely higher-accuracy replacements for correlations. They also clarify variable importance, expose extrapolation limits, and test whether a model is interpolating within a dense database or predicting outside the training envelope. For benchmark design, 0+0D datasets should include conventional correlation baselines, uncertainty, fluid- or facility-held-out splits, and enough metadata to compute nondimensional groups.

\subsection{ML methods for 0+1D time-series data}

The 0+1D class includes thermocouple histories, pressure traces, hydrophone waveforms, acoustic-emission hits, microphone recordings, and power histories. These signals are well suited to spectral features, time-frequency representations, recurrent networks, temporal convolutional networks, transformers, Gaussian-process models, and sequence-regression workflows [118-119]. Acoustic sensing is especially important because it can operate when optical access is limited. Recent acoustic boiling studies demonstrated acoustic-emission spectrograms for boiling-regime and boiling-crisis prediction, released associated audio and code, and later framed acoustic boiling diagnostics as a broader roadmap for regime identification and crisis warning [14-16,23,120].

NED3 acoustic datasets extend this direction toward quantitative heat-flux regression and multimodal fusion. Hydrophone data, AE hits, and waveforms have been paired with temperature histories, heat-flux labels, and bubble images for transient and steady-state boiling analysis [17-18,24-26,52]. The main methodological risk in this S+TD class is leakage across time windows from the same run. Benchmark splits should therefore be defined at the run, surface, fluid, or facility level, not by randomly sampling short windows from a long waveform.

\subsection{ML methods for 1+0D and 1+1D profiles}

The 1+0D and 1+1D classes receive less attention than images and waveforms, but they are important bridges between point sensors and full fields. Examples include line-scan temperature profiles, pressure distributions, centerline velocity profiles, spatial heat-flux profiles, and time-evolving interface or temperature profiles. Suitable models include profile regression, principal-component analysis, dynamic mode decomposition, Gaussian-process regression, temporal sequence-to-sequence models, and reduced-order representations. These approaches are especially useful when an experiment cannot support full-field imaging but can still resolve a spatial coordinate.

For two-phase benchmarks, profile data should preserve the coordinate system, sensor spacing, interpolation method, boundary conditions, and uncertainty. A 1+1D wall-temperature profile over time can also support inverse heat-conduction and physics-discovery tasks when the raw measurements and sensor geometry are retained. This point connects directly to uncertainty-aware inverse modeling and governing-equation discovery, where raw histories and spatial sensor layout are as important as the final reconstructed heat flux.

\subsection{ML methods for 2+0D images}

The 2+0D class includes single optical images, infrared thermographs, surface micrographs, phase maps, and simulation slices. These data are naturally handled by convolutional neural networks, vision transformers, object detectors, segmentation models, and image-regression models [121-128]. Early boiling-visualization studies showed that even low-speed or low-resolution images could support regime classification and heat-flux estimation using support-vector machines, neural networks, PCA, and compact image features [13,75]. Suh et al. connected high-speed bubble images with boiling-curve prediction and showed that visual bubble statistics can act as an interpretable bridge between interface dynamics and thermal performance [76].

Open-source image-analysis tools reinforce the value of this S+TD class. BubMask and BubbleID use modern segmentation and tracking workflows to extract bubble masks, centroids, areas, axes, interface information, vapor fraction, and departure-related features [29-30,79-81]. These tools are most useful when paired with annotated data, because segmentation labels, bubble definitions, and interface masks can differ across groups. Benchmark image datasets should therefore report pixel scale, field of view, lighting, annotation protocol, surface condition, and train-test splits that separate experimental runs rather than adjacent frames.

\subsection{ML methods for 2+1D videos and infrared sequences}

The 2+1D class includes high-speed boiling videos, IR videos, event-camera streams, and time-resolved two-dimensional simulation slices. These records enable video classification, segmentation over time, optical-flow-like analysis, sequence regression, event prediction, dryout detection, and optical-to-thermal inference. Broader multimodal diagnostics have shown how optical imaging, infrared fields, acoustic and acoustic-emission sensing, and event streams can capture complementary projections of boiling and two-phase-flow behavior [14-18,23,82-86,129-130]. High-speed IR thermometry has been especially powerful for boiling crisis because it resolves the thermal footprint of liquid-solid contact, dryout, and heat-transfer nonuniformity. High-resolution IR studies have used neural networks to infer bubble statistics, predict distance to boiling crisis, analyze dry-area evolution, and support mechanistic boiling-crisis criteria [8-13]. For condensation, Khodakarami et al. combined optical images, infrared measurements, and generative learning to map visible droplet dynamics into local temperature fields, illustrating how cross-modal models can reduce the experimental burden of thermal-field measurement when calibration and physical conditioning are retained [85-86].

Optical 2+1D data provide a complementary view of the interface. Recent flow-boiling visualization work used machine-learning vision tools to extract void fraction, vapor-liquid interfacial behavior, and wall wetting-front areas from flow-visualization data and then used those quantities to support void-fraction, HTC, and CHF modeling [77]. VISION-iT provides a modular video-analysis pipeline for acquisition, object detection, tracking, and postprocessing of bubble and droplet features [32]. Bubble2Heat uses segmented high-speed recordings and pointwise temperature information for optical-to-thermal inference [21], while EventFlow uses event-based sensing for low-latency boiling-flow-regime classification [22]. These examples show why 2+1D data need explicit frame rate, exposure, synchronization, heat-load protocol, and video-level splits.

\subsection{ML methods for 3+0D, 3+1D, and simulation-derived fields}

The 3+0D and 3+1D classes include steady or transient CFD fields, volumetric reconstructions, holographic fields, and digital-twin state histories. Scientific machine learning methods for these data include reduced-order models, autoencoders, graph neural networks, neural operators, differentiable surrogates, and physics-informed neural networks [131-141]. Broader reviews of machine learning for fluid mechanics and scientific-knowledge integration provide useful context for why these methods need physics-aware features, constraints, and evaluation metrics rather than only generic prediction loss [142-143]. Simulation datasets are valuable because they can provide dense pressure, velocity, temperature, and phase fields that are difficult to measure experimentally. BubbleML is a prominent example because it provides physics-driven boiling simulations and benchmark tasks for optical flow, operator learning, and spatiotemporal prediction [19,35]. Bubbleformer and BubbleML 2.0 extend this direction from short-range prediction toward autonomous long-range forecasting of boiling dynamics, including nucleation, interface evolution, velocity fields, and heat transfer across fluids, geometries, flow regimes, and boundary conditions [40-43]. NUCLEUS-MoE further illustrates the move toward unified simulation surrogates by combining mixture-of-experts routing, neighborhood attention, and interface-consistency treatment to model saturated and subcooled pool boiling across dielectrics, refrigerants, cryogens, and immersion-cooling fluids [49-50]. HB-ARFM shifts the task from forward forecasting to inverse reconstruction by using interface observations extracted from high-speed imaging to infer hidden velocity and temperature fields through history-bootstrapped autoregressive flow matching [46-47]. FD-Bench provides another signal that two-phase boiling is entering general SciML benchmarking, because it evaluates multiple operator and neural-simulation model families on BubbleML pool- and flow-boiling cases [48]. These expansions are important for the S+TD taxonomy because they treat simulation-derived boiling records as time-resolved field data with thermophysical conditioning, partial-observation operators, model-family comparisons, and physics-based evaluation metrics, rather than only as image sequences for generic video prediction.

Simulation-oriented bubble-dynamics repositories provide another part of the landscape. BubbleNet applies physics-informed deep learning to bubble flow in microfluids [36], while Rayleigh-Plesset and Keller-Miksis repositories support reduced-order bubble dynamics and surrogate modeling [37-39]. The limitation is that simulation fields are not experimental truth. They depend on closure assumptions, boundary conditions, mesh resolution, material properties, and numerical implementation. A useful simulation benchmark should therefore include solver details, boundary conditions, mesh/time-step information, validation or sanity checks, and test cases that examine transfer to experimental data.

\subsection{ML methods for mixed S+TD multimodal datasets}

Mixed S+TD datasets combine synchronized measurements from multiple classes, such as optical videos, IR fields, acoustic waveforms, AE hits, thermocouple histories, pressure traces, and heat-flux labels. These datasets are especially valuable because each modality observes only part of the coupled two-phase process. A boiling video may show bubble footprints but not subsurface temperature; an IR image may show dryout and rewetting but not all interface geometry; an acoustic waveform may integrate many bubble events but lose spatial localization; and a CFD field may provide complete state variables but depend on closure assumptions.

Multimodal learning therefore supports cross-modal prediction, sensor fusion, missing-modality robustness, self-supervised pretraining, and experiment-simulation transfer. The key requirement is synchronization metadata. A dataset that aligns acoustic signals, optical videos, IR fields, heat-flux labels, and operating conditions can test whether sound predicts optical regime changes, whether images infer local heat flux, whether simulation-trained representations transfer to laboratory measurements, and whether a model remains accurate when one modality is removed.

\subsection{Sequence-regression workflows for mixed S+TD data}

Sequence-regression workflows provide a useful bridge between heterogeneous thermal-fluid measurements and ML models. In boiling heat-transfer experiments, inputs may be hydrophone waveforms, AE hit sequences, optical-image features, or short video windows, while outputs may be heat flux, temperature, regime labels, or transition indicators. This structure appears repeatedly across S+TD classes because the input and output are rarely sampled in the same way.

SeqReg is one open-source example of this workflow. The repository defines a pipeline for loading experimental datasets, converting long records into shorter input-output sequences, optionally transforming signals into the frequency domain, defining or loading a model, training or evaluating the model, and analyzing prediction performance [87]. The reported workflows include hydrophone-based, AE-hit-based, and optical-image-based heat-flux prediction tasks [87]. Similar workflow abstractions will be needed for future benchmarks because they make preprocessing, windowing, splitting, and evaluation rules explicit rather than hidden inside single-use scripts.

\begin{figure}[H]
\centering
\includegraphics[width=0.95\linewidth]{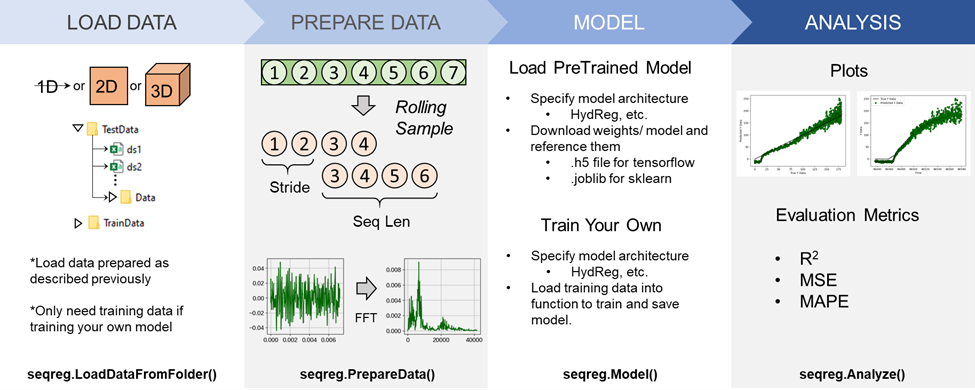}
\caption{SeqReg workflow from loading data to preparing sequences, defining or loading models, and analyzing prediction performance.}
\label{fig:seqreg_workflow}
\end{figure}

\begin{figure}[H]
\centering
\includegraphics[width=0.95\linewidth]{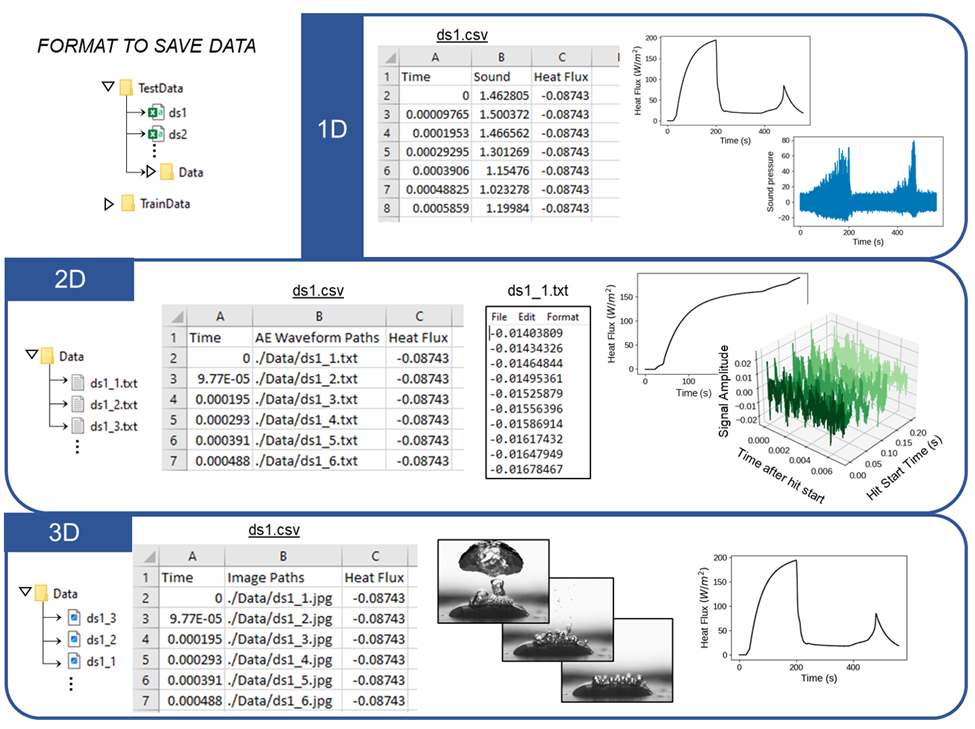}
\caption{Multidimensional sequential data used by SeqReg, illustrating a data-format strategy for 1D, 2D, and 3D input types and connecting SeqReg to the S+TD framework for hydrophone, AE, image, and video datasets.}
\label{fig:seqreg_data}
\end{figure}

Figure~\ref{fig:seqreg_workflow} summarizes the sequence-regression workflow from data loading, to data preparation, model loading/training, and performance evaluation. Figure~\ref{fig:seqreg_data} shows why the same logic can support several S+TD classes, connecting 0+1D acoustic signals, 2+1D image sequences, and mixed synchronized data to scalar or sequence outputs such as heat flux. The broader take-home message is not specific to one package. Benchmark datasets should release the data-preparation logic together with the raw records, because model performance can change substantially with window length, frequency transforms, synchronization choices, and run-level splitting.

\subsection{Generalization checks and leakage risks}

Across all S+TD classes, the most important unresolved ML challenge is generalization across domains. A model trained on one heater, surface, fluid, pressure range, optical configuration, sensor placement, or simulation setup may perform well under random splitting and fail under laboratory-held-out or surface-held-out evaluation. This problem is especially severe for 2+1D videos and 0+1D waveforms because adjacent frames or windows from the same experiment are highly correlated. It also appears in tabular databases when dimensional variables allow the model to memorize fluid or facility identity rather than learning transferable physics.

The benchmark implication is straightforward. Two-phase AI should report not only accuracy, but also the split logic. Useful splits include run-held-out, surface-held-out, fluid-held-out, geometry-held-out, heat-load-path-held-out, and laboratory-held-out evaluations. Models should also be checked for physical consistency, such as impossible heat fluxes, nonphysical monotonicity, invalid regime transitions, and sensitivity to calibration or filtering choices. This challenge subsection is intentionally placed after the S+TD method review because the leakage mode depends on the data class. A bad split in a waveform benchmark is not the same as a bad split in a CFD field or a tabular correlation database.

\begin{figure}[H]
    \centering
    \includegraphics[width=1\linewidth]{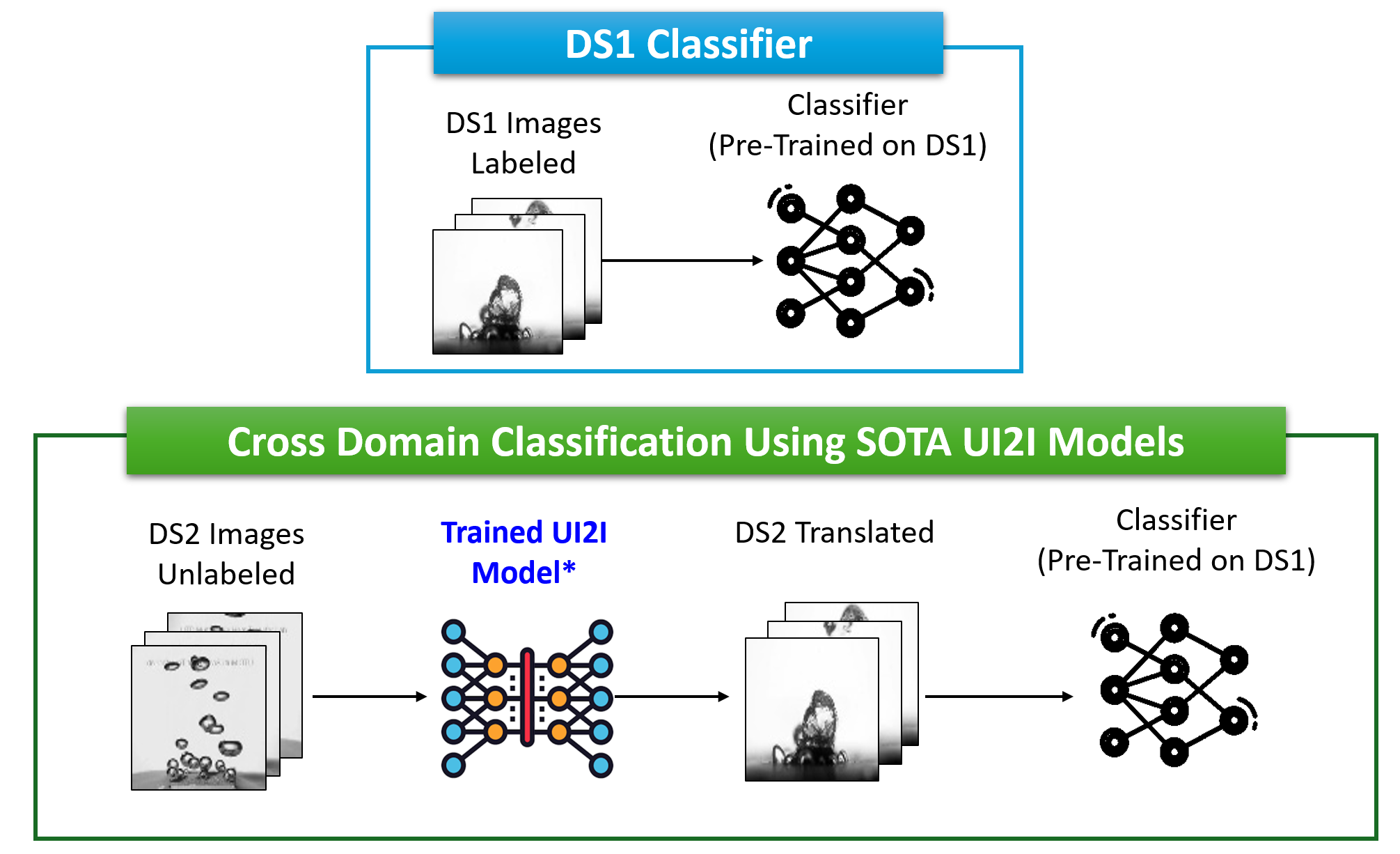}
    \caption{Cross-domain classification through unsupervised image-to-image translation. The upper panel shows a classifier trained on labeled source-domain data (DS1). The lower panel illustrates the domain adaptation workflow, where unlabeled target-domain images (DS2) are translated into the source-domain representation using a trained image-to-image translation model. By aligning the visual distributions of the two domains, the translated images can be classified using the existing DS1 model, mitigating performance degradation caused by domain shift while eliminating the need for additional target-domain annotations. Reproduced with permission from [145], License No. 6282770928729.}
    \label{fig:UI2I_Solution}
\end{figure}

Figure~\ref{fig:UI2I_Solution} illustrates one representative response to this challenge for 2+0D boiling images. Instead of retraining the classifier for every new facility, a target-domain image can be translated into the source-domain representation on which the classifier was trained. This strategy does not remove the need for benchmark discipline, but it shows why future datasets should include domain-shift scenarios rather than only random train-test splits [144-148].

The following subsection discusses these image-translation methods as examples of physics-aware generalization for image-based thermal-fluid AI.

\subsection{Cross-domain generalization for 2+0D image datasets}

To facilitate the development of robust AI models for thermal-fluid diagnostics, several methods have been developed to address cross-domain generalization, one of the most persistent challenges in the field. Thermal-fluid 2+0D image datasets are often collected using different experimental facilities, heater geometries, working fluids, operating conditions, imaging systems, and sensor configurations. As a result, machine learning models trained on one dataset frequently experience significant performance degradation when applied to data acquired under different conditions. Earlier work [147] addressed cross-domain generalization in 2+0D boiling-image datasets using transfer learning, in which a convolutional neural network trained on a source dataset was adapted to a new domain using a limited set of labeled target-domain samples. While this approach reduced annotation requirements, it still relied on the availability of labeled data from the target domain, motivating the development of fully unsupervised adaptation methods that eliminate the need for additional annotation altogether.

\begin{figure}[H]
    \centering
    \includegraphics[width=1\linewidth]{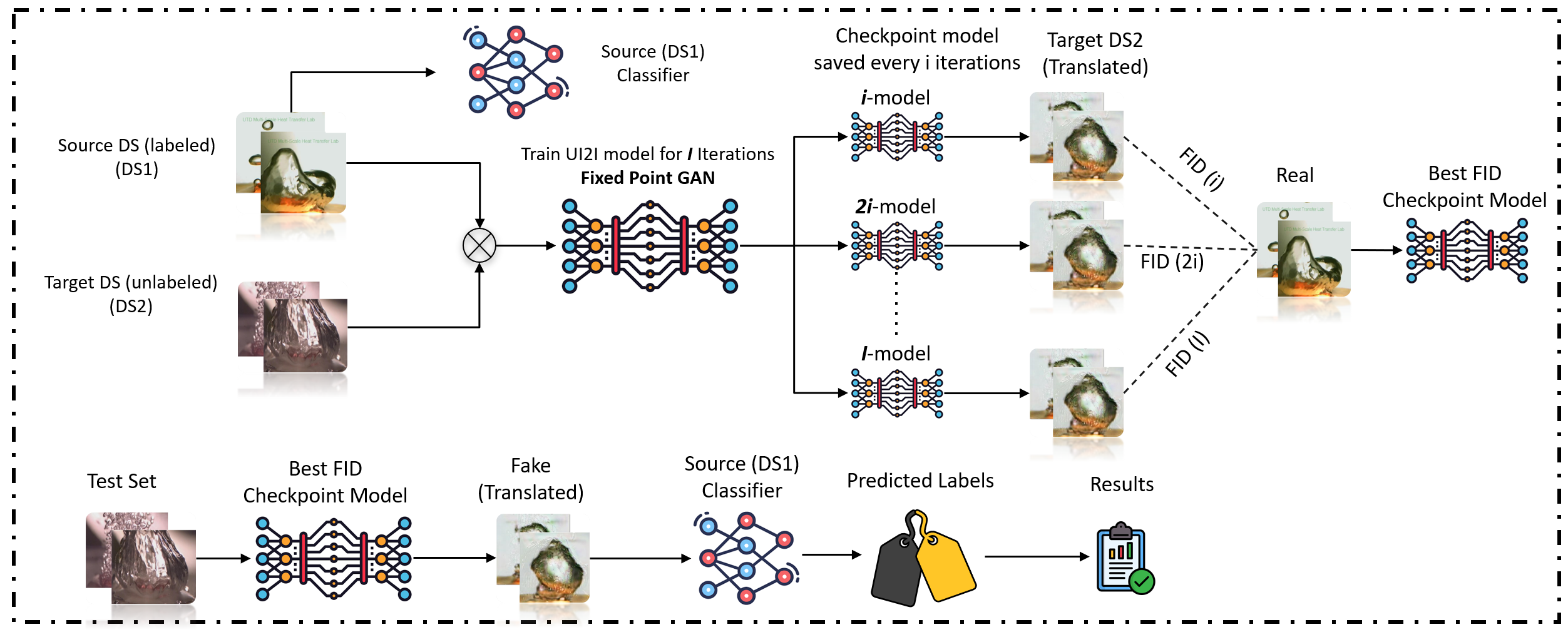}
    \caption{Overview of the UI2I framework introduced by Al-Hindawi et al. [144] for 2+0D domain adaptation. An unpaired image-to-image translation model is trained using labeled source-domain images (DS1) and unlabeled target-domain images (DS2). Model checkpoints are evaluated using the Frechet Inception Distance (FID), and the checkpoint with the lowest FID is selected. During inference, target-domain images are translated into the source-domain style and subsequently classified using the pre-trained source-domain classifier without requiring target-domain labels.}
    \label{fig:UI2I_Framework}
\end{figure}

To eliminate the dependence on labeled target-domain data, subsequent studies explored fully unsupervised domain adaptation using unsupervised image-to-image translation (UI2I). The work by Al-Hindawi et al. [144-146,148] leveraged UI2I translation to address domain shift in boiling-crisis detection. As shown in Figure~\ref{fig:UI2I_Framework}, the framework decouples domain adaptation from the classification task by introducing a preprocessing stage that translates target-domain images into the visual style of the source domain before classification. Rather than adapting the classifier itself, the translated images are processed using a previously trained classifier without requiring retraining, fine-tuning, or additional annotation effort. This strategy showcased the portability of boiling-crisis detection models across experimental facilities and operating conditions while maintaining a fully unsupervised workflow. 

\begin{figure}[H]
    \centering
    \includegraphics[width=1\linewidth]{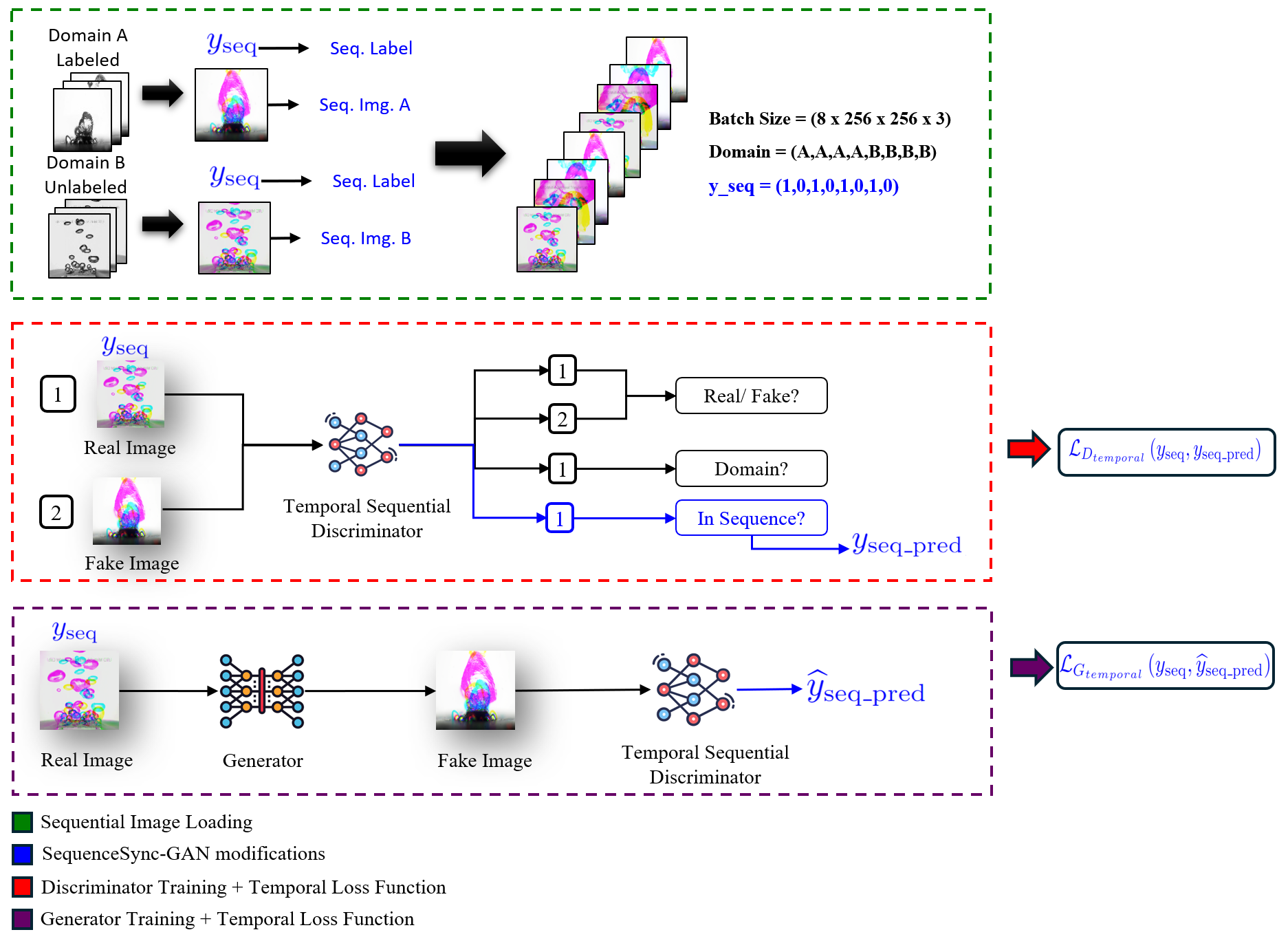}
    \caption{SequenceSync-GAN architecture for sequence-aware domain adaptation. The framework augments conventional image-to-image translation with temporal sequence labels, a Temporal Sequential Discriminator, and temporal consistency loss functions for both discriminator and generator training. These additions aim to softly encourage translated image sequences to preserve the temporal evolution of boiling dynamics during translation, resulting in improved cross-domain boiling crisis detection performance. Adapted with permission from [145], License No. 6282770928729.}
    \label{fig:SequenceSync_framework}
\end{figure}

In a more recent work, SequenceSync-GAN [145] improved upon prior work by softly incorporating temporal information directly into the UI2I translation model to enhance domain adaptation. As shown in the framework in Figure~\ref{fig:SequenceSync_framework}, rather than treating each image independently, the framework introduces temporal data-loading strategies, sequence-aware discriminators, and temporal consistency losses that preserve the chronological progression of boiling phenomena during domain translation. This sequence-aware approach improved cross-domain classification performance while highlighting the importance of incorporating process dynamics into domain adaptation workflows.

\begin{figure}[H]
    \centering
    \includegraphics[width=1\linewidth]{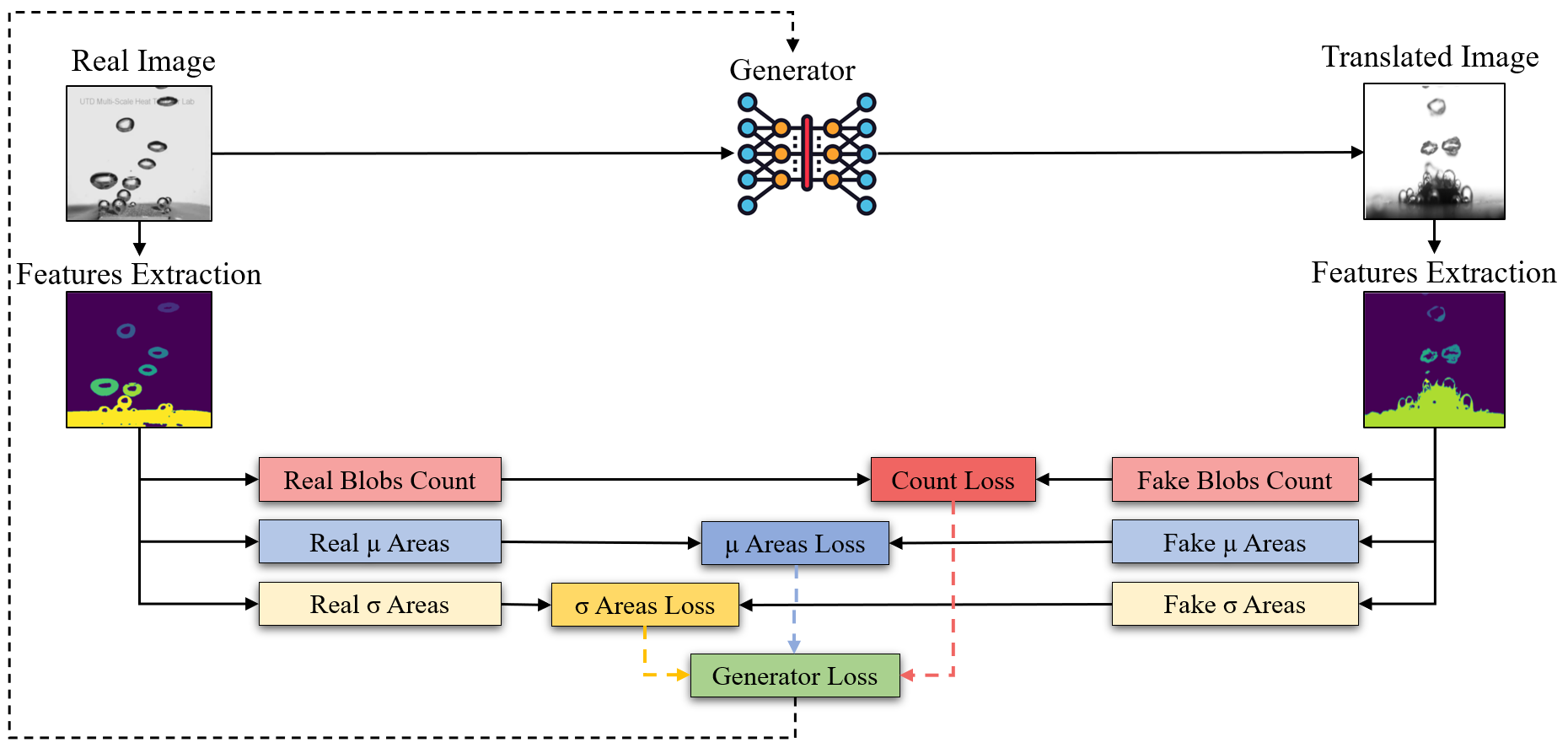}
    \caption{BubbleSync-GAN framework for physics-informed domain adaptation. Bubble-level features are extracted from both real and translated images through unsupervised segmentation, and feature-preservation losses based on bubble count, mean bubble area ($\mu$), and area standard deviation ($\sigma$) are incorporated into generator training. By constraining the translation process using physically meaningful boiling descriptors, the framework improves cross-domain adaptation while preserving the underlying boiling physics.}
    \label{fig:BubbleSync_Framework}
\end{figure}

Although image-to-image translation can improve domain alignment, unconstrained translation may inadvertently alter physically meaningful flow characteristics. The BubbleSync-GAN framework [146], shown in Figure~\ref{fig:BubbleSync_Framework}, addresses this challenge through physics-informed domain adaptation. The framework extracts bubble-level descriptors and incorporates physics-guided constraints that preserve key boiling characteristics, including bubble count, average bubble size, and bubble-size distributions during translation. By enforcing consistency in physically relevant features, BubbleSync-GAN improves domain adaptation performance while maintaining fidelity to the underlying thermal-fluid physics.

These studies illustrate the evolution of cross-domain thermal-fluid AI from supervised transfer learning to fully unsupervised domain adaptation and, more recently, toward physics-informed and temporally informed adaptation strategies. The progression highlights an important lesson for future benchmark development. Successful thermal-fluid AI models must generalize across facilities, operating conditions, and sensing modalities rather than relying solely on performance within a single experimental dataset. Consequently, future benchmark datasets should explicitly include cross-domain evaluation protocols and domain-shift scenarios as part of their evaluation framework.

\subsection{Machine learning for inverse heat conduction in spatiotemporal domains}
\label{subsec:ihcp}

Inverse heat conduction problems (IHCPs) arise when the goal is to infer unknown thermal quantities, most commonly the boundary heat flux, from indirect interior
temperature measurements rather than from prescribed forcing conditions. Unlike the
well-posed forward problem, the IHCP is mathematically ill-posed, meaning that small
perturbations in measured temperatures can produce large, unstable changes in the
recovered heat flux and make direct inversion impractical [149-150]. This challenge is
particularly relevant to two-phase heat-transfer experiments, where transient boiling
events, critical-heat-flux onset, and phase-transition dynamics produce rapidly evolving,
spatially non-uniform temperature fields that must be reconstructed from a finite set of
thermocouple readings distributed inside the heater block.

Classical regularization and sequential function specification approaches [149-150] and early
Bayesian formulations [151] address ill-posedness but treat the measured temperature
field as either a spatially averaged or temporally independent signal. In practice,
thermocouples sample temperature simultaneously at several fixed locations over many
time steps, so the data carry both spatial and temporal structure. Ignoring the
interaction between these two dimensions discards information that is particularly
valuable during transient boiling events, where the temperature field evolves
non-uniformly across the heater surface. Physics-informed neural networks have also been
applied to forward and inverse heat-conduction problems [132,152], but these approaches
similarly do not exploit the joint spatiotemporal covariance structure inherent in
multi-sensor measurements.

\begin{figure}[H]
    \centering
    \includegraphics[width=0.85\linewidth]{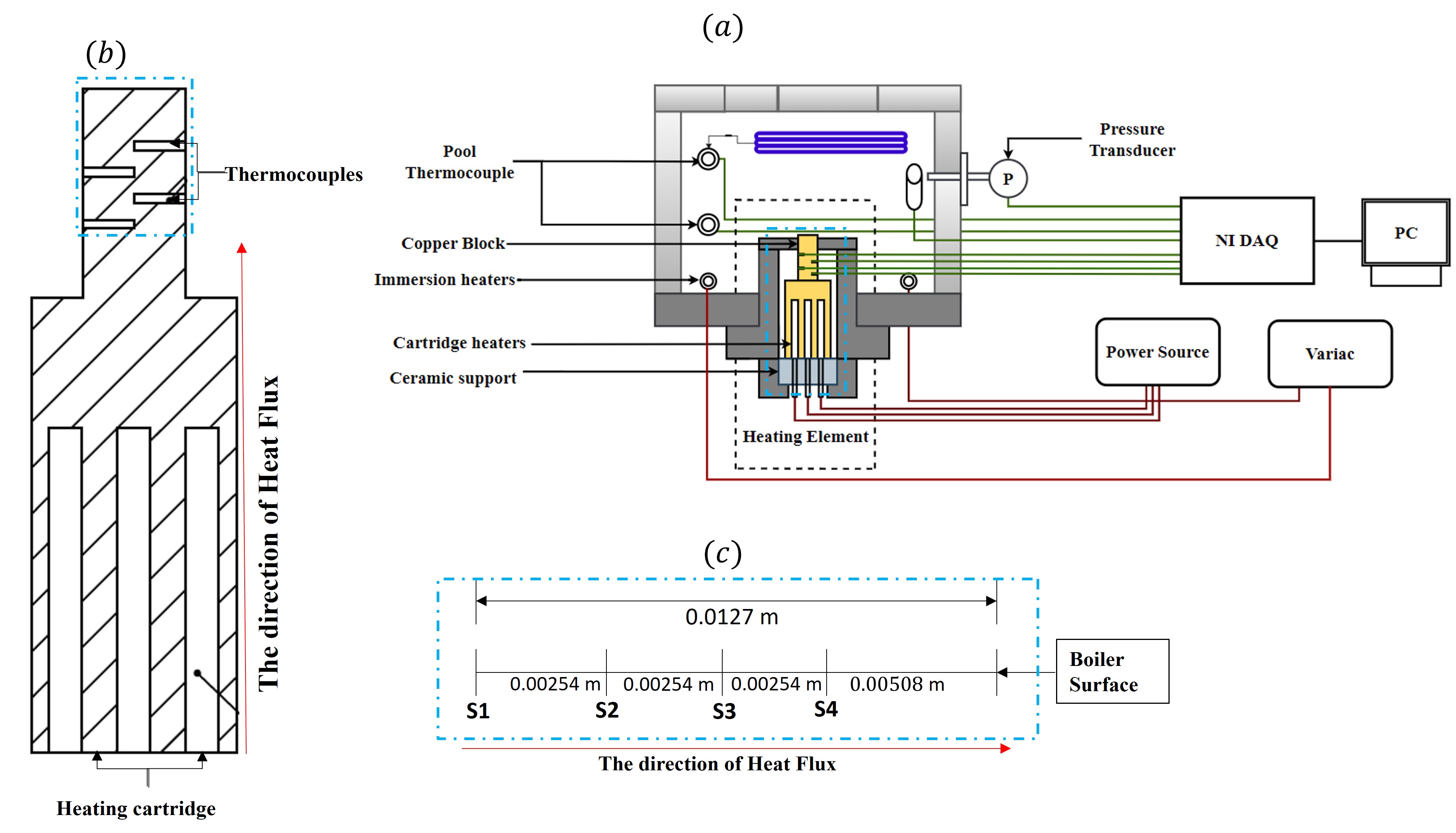}
    \\[8pt]
    \includegraphics[width=0.85\linewidth]{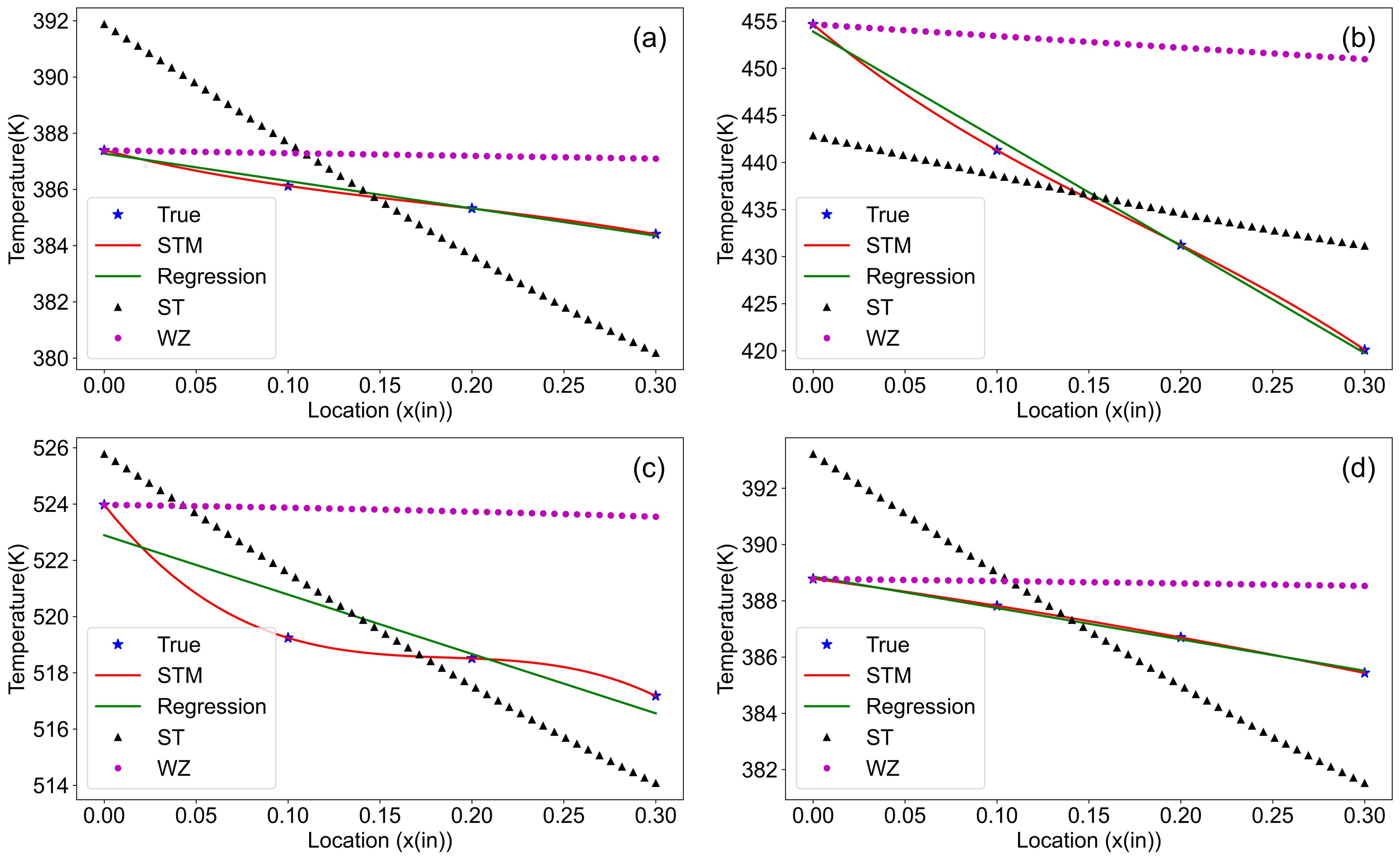}
    \caption{Pool boiling experiment and Bayesian spatiotemporal model (STM) results
    for inverse heat-flux reconstruction. \textit{Top}: experimental setup showing
    $(a)$~the pool boiling apparatus, $(b)$~details of the heating element, and
    $(c)$~1D schematics of the thermocouple region (S1--S4). \textit{Bottom}: spatial
    temperature predictions at times $(a)$~100\,s, $(b)$~176\,s, $(c)$~200\,s, and
    $(d)$~400\,s, comparing STM, the interaction-free baseline (ST), the Wang--Zabaras
    method (WZ), and linear regression against thermocouple measurements; STM achieves
    substantially lower RMSE than all baselines in both space and time. Adapted
    from~[153].}
    \label{fig:stm_combined}
\end{figure}

Olabiyi et al. addressed this gap with a Bayesian spatiotemporal
modeling (STM) approach that explicitly incorporates spatial, temporal, and
spatial-temporal interaction terms into a penalized-spline forward model, with Gaussian
Markov Random Field priors placed on the corresponding coefficient sets [153]. A Gibbs
sampling algorithm draws from the joint posterior, yielding not only a point estimate of
the unknown heat flux but also quantified uncertainty, which is essential for propagating
reconstruction error into downstream boiling-curve analysis or CHF detection. Evaluated
on a pool-boiling experiment with a copper heater block instrumented with four T-type
thermocouples (Figure~\ref{fig:stm_combined}), the STM achieved substantially lower
spatial and temporal RMSE than the Wang--Zabaras Bayesian model [151], an
interaction-free spatiotemporal baseline, and a linear regression approach.

In the S+TD framework, the pool-boiling thermocouple dataset used in this study is a
1+1D spatiotemporal record, where the four thermocouple locations define the spatial
coordinate and time defines the temporal coordinate. The STM therefore bridges the
0+1D acoustic and 2+1D video methods reviewed elsewhere in this section with a rigorous
statistical treatment of the spatiotemporal structure inherent in multi-sensor heat-flux
reconstruction. For benchmark purposes, this work underscores that raw temperature
histories, sensor coordinates, calibration details, and repeated-condition variability
should be preserved in open datasets and not only the final processed heat-flux labels,
so that future users can test both predictive ML and physics-based inverse methods.

\subsection{Machine learning for discovering governing equations}
\label{subsec:physics_discovery}

A complementary challenge to inverse heat-flux reconstruction is the identification of governing equations directly from experimental data. In complex thermal-fluid systems, the correct form of the governing physics is not always known a priori, and this uncertainty motivates data-driven approaches to equation discovery. The Sparse Identification of Nonlinear Dynamics (SINDy) framework [131] formulates this as a sparse regression problem in which observed state trajectories are regressed onto a library of candidate physics terms and sparsity-promoting penalties select the active subset. Extensions such as Ensemble-SINDy [154], Weak-SINDy [155], and Bayesian variants [156-157] have since improved robustness to measurement noise and limited data. Despite these advances, a critical limitation persists across existing methods. Governing equation coefficients are treated as fixed quantities whose uncertainty arises solely
from measurement noise. Real physical systems, however, frequently exhibit genuine
variability in their governing parameters arising, for example, from variability in
fluid viscosity, surface roughness, or operating pressure. Ignoring this input variability conflates two fundamentally different sources of uncertainty, leading to biased coefficient estimates and unreliable structural identification. This is
particularly consequential in thermal-fluid systems, where material properties and
operating conditions vary across experiments and across laboratories, and where the
forms of closure terms and interfacial momentum balances are themselves uncertain.

\begin{figure}[H]
    \centering
    \includegraphics[width=0.95\linewidth]{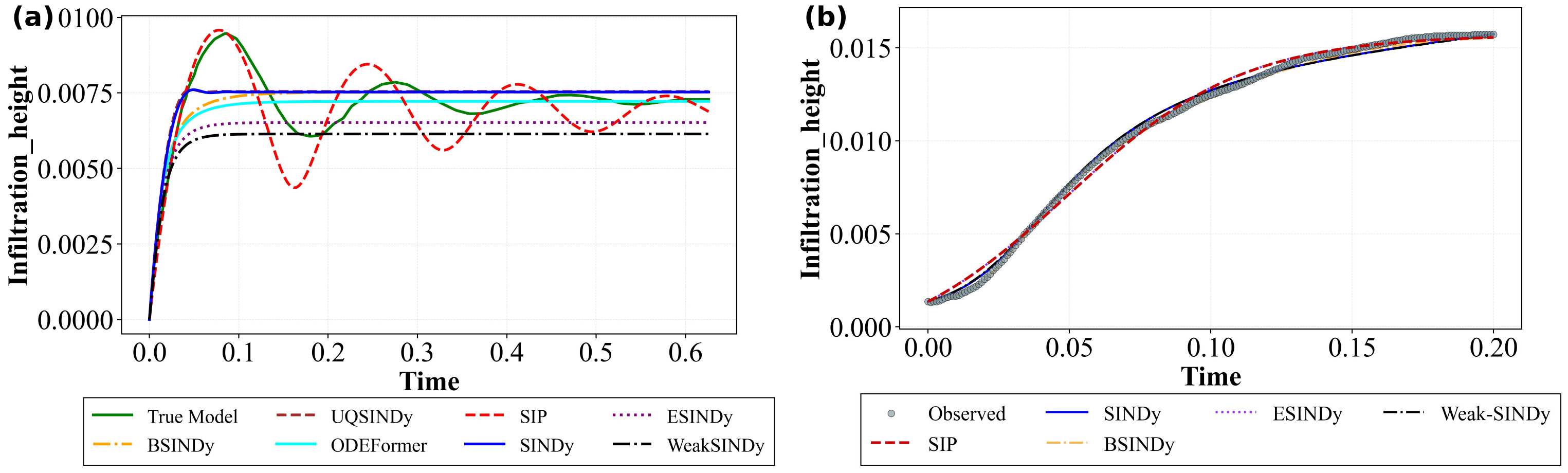}
    \caption{SIP equation-discovery results for capillary infiltration in porous media. $(a)$~Low-viscosity fluid (ether): SIP correctly captures the inertia-driven oscillatory transient before equilibrium, while competing methods introduce spurious cubic or quartic polynomial terms that misrepresent the underlying physics. $(b)$~High-viscosity fluid (silicone oil): all methods achieve comparable trajectory fits, but only SIP achieves perfect structural identification (DiCE\,$= 1.0$), recovering all four physical terms of the momentum balance without spurious
    additions. Adapted from~[158].}
    \label{fig:sip_infiltration}
\end{figure}

Olabiyi et al. addressed this gap by introducing the stochastic inverse physics-discovery (SIP) framework [158], which re-envisions the unknown governing-equation coefficients as random variables rather than fixed scalars. SIP infers their posterior distributions by minimizing the Kullback--Leibler divergence between the push-forward of posterior samples through the candidate governing equation and the empirical distribution of the observed data, simultaneously accounting for both intrinsic input variability and measurement noise. Sparsity-promoting priors enforce parsimony, and a matched-block bootstrap procedure [159] generates multiple sample paths when only a single experimental trajectory is available, which is a common constraint in boiling and infiltration experiments. Applied to a capillary infiltration problem [160] that involves a four-term momentum balance coupling inertial, viscous, gravitational, and capillary forces, SIP was the only method among eight state-of-the-art comparators to achieve perfect structural identification (DiCE\,$= 1.0$) for both low- and high-viscosity fluids, correctly recovering all four physical terms without spurious additions. Figure~\ref{fig:sip_infiltration} shows the predicted infiltration height trajectories for both fluid systems, illustrating how SIP correctly captures the inertia-driven oscillatory transient for ether and the monotonic capillary-driven rise for silicone oil, while competing methods introduce spurious polynomial terms or miss essential physics entirely. Compared with methods spanning sparse regression (SINDy [131], Ensemble-SINDy [154], Weak-SINDy [155]), probabilistic sparse regression (UQ-SINDy [156], B-SINDy [157]), and symbolic regression, SIP reduced coefficient RMSE by approximately 82\% on average across ten test scenarios and provided well-calibrated 95\% credible intervals (94.2\% empirical coverage).

For two-phase heat-transfer benchmarks, this framework suggests that open datasets should preserve raw temperature histories, sensor geometry, repeated-condition variability, and known material properties, and not only the final processed heat-flux labels. With these ingredients available, SIP-style physics-discovery methods can be tested for their ability to identify governing constitutive relations, interfacial momentum balances, or closure terms, which are tasks that go beyond scalar regression and connect open data directly to the governing equations of multiphase transport. Together with the STM approach reviewed above, these Bayesian statistical methods show how models that respect the spatiotemporal and stochastic structure of thermal-fluid experiments can advance both parameter estimation and equation discovery in support of physics-interpretable, open-data machine learning.

\subsection{Broader model families and benchmark analogues}

The take-home message from the broader literature is that two-phase AI is moving from isolated demonstration models toward reusable task families. Consolidated-data studies on flow-boiling HTC, flow-condensation HTC, pressure-drop prediction, and mixture condensation show how tabular ML can evolve into database infrastructure for phase-change correlations [161-164]. Synchronized optical and infrared boiling studies show a different path, where aligned phase maps, temperature fields, and inferred heat-flux fields create targets for physics-aware computer vision and thermal-field learning [165-166]. Image-based, reduced-order, clustering, and geometry-conditioned surrogate studies further expand the field beyond supervised scalar prediction by extracting dynamic descriptors, building nonintrusive moving-boundary surrogates, identifying interpretable regimes, and learning image-to-field mappings [27-28,167-168].

Older AI predictors and recent CHF studies show that data-driven boiling and CHF modeling predates the current deep-learning wave and now spans fuzzy models, neural-network CHF predictors, physics-informed prediction, benchmark-style model assessment, surface-structure-specific learning, and deep-learning CHF detection [147,169-173]. A separate image-analysis cluster includes event-based boiling-crisis prediction, BubbleMask-style visualization, U-Net and optical-flow bubble analysis, and pressure-dependent bubble segmentation [174-176]. Other studies broaden the review from images to dynamical systems and regime maps through RQA/PCA flow-pattern identification, cluster-based reduced-order descriptions from tomography, annular-flow regime identification, CNN/Bayesian film-boiling detection, and ML-enabled condensation heat-transfer measurement [33-34,177-181]. Synchronized optical and infrared measurements on surfaces with different dynamic wettability reinforce the need to release raw signals, processed fields, labels, and physical metadata together rather than only final HTC or CHF values [166].

Benchmark examples from neighboring fields clarify what the two-phase community should borrow and what it should avoid. ImageNet illustrates the power of a large curated corpus with fixed recognition tasks and common evaluation rules [182]. COCO is a closer analogue for multimodal boiling data because one corpus supports multiple tasks, including detection, segmentation, and captioning [183]. Food-101 and FoodSeg103 show how focused domain datasets can become useful benchmarks when they provide clear classes, fixed splits, and pixel-level labels [184-185]. Robomimic is relevant because it couples datasets, learning algorithms, examples, and reproducible workflows for offline learning from demonstrations [186]. FD-Bench adds a closer scientific-ML analogue by comparing model families across fluid-simulation tasks and including two-phase boiling cases from BubbleML [48]. For two-phase heat transfer, these analogues point to a practical rule. A useful benchmark should include data loaders, baseline scripts, task definitions, split files, metadata schemas, evaluation metrics, benchmark reports, and versioned community contributions.

The S+TD taxonomy also connects boiling-focused work to adjacent heat-transfer and multiphase-flow AI studies. Broad heat-transfer ML reviews emphasize that the field is not limited to deep-learning image analysis, since reduced-order modeling, coefficient and pressure-drop prediction, real-time experiment analysis, and system-level optimization are recurring use cases across thermal science [187-188]. Correlation-learning studies remain important because many benchmark labels will still be scalar coefficients, pressure drops, Nusselt numbers, or CHF margins rather than images or fields [189]. Interpretable architectures such as DimNet for microfin-tube flow boiling are especially useful because they can be converted into explicit piecewise power-law-like functions and therefore sit closer to traditional dimensionless correlations than to black-box regression [190]. Cryogenic examples, including ANN-based saturated hydrogen nucleate flow-boiling prediction, show that data-driven heat-transfer models are also valuable in regimes where fluid properties and operating constraints differ sharply from water or refrigerant benchmarks [191].

Two-phase flow-pattern and void-fraction studies outside pool boiling broaden the same S+TD framework. Reviews of non-invasive measurement and control highlight tomography, signal processing, and control-oriented diagnostics as important precursors to data-driven sensing [192]. Studies using the Shoham inclined-pipe database show that public or semi-public flow-regime databases can support standardized classification tasks even when the raw diagnostics differ from high-speed boiling videos [193]. Recent image-processing and ML work on void-fraction estimation in corrugated channels further bridges computer vision, multiphase-flow reconstruction, and physically interpretable scalar outputs [194]. Condensation and surface-engineering studies add another generalization constraint. Micro/nanoengineered boiling and condensation surfaces are controlled by morphology, chemistry, wettability, manufacturability, durability, and pressure-drop tradeoffs [195]. Vision-based condensation learning connects droplet-level image statistics to heat-transfer performance [181]. These adjacent literatures support a central recommendation of this review. Benchmarks should be organized by transport mechanism, diagnostic modality, physical metadata, operating regime, and intended generalization test, not only by model architecture.

\subsection{Roadmap for two-phase heat-transfer datasets and machine-learning models}

The S+TD review points to an infrastructure problem as much as a modeling problem. The FAIR principles provide the general standard that data should be findable, accessible, interoperable, and reusable [57]. Research-data repositories, data-citation practices, open-source repository software, persistent identifiers, and machine-readable metadata schemas provide the institutional layer needed for durable datasets [58-64]. Engineering datasets also need format standards. Self-describing formats such as HDF5 and NetCDF are useful for large multidimensional arrays, but they do not by themselves define the physical meaning of variables, units, coordinate systems, synchronization, calibration, uncertainty, or valid operating range [65-67]. For two-phase heat transfer, FAIR data must therefore become physics-FAIR data.

Benchmark governance is the second infrastructure layer. Machine-learning reproducibility studies show that leakage, undocumented preprocessing, unavailable code, hidden hyperparameter choices, and ambiguous train-test splits can make reported performance difficult to trust [44-45,68-71]. Two-phase heat transfer is especially vulnerable because adjacent frames from a single video, neighboring operating points from a single boiling curve, or repeated tests from a single surface can look independent to a generic ML workflow while remaining physically coupled. Good benchmarks should include predefined splits, baseline models, raw-to-processed data provenance, uncertainty estimates, failure cases, and a written rule for whether new submissions may use external data.

Table~\ref{tab:4} condenses the review into a benchmark-oriented design matrix. It shows that two-phase AI is not a single modeling problem. Scalar correlations, image segmentation, acoustic sequence learning, IR inverse analysis, simulation surrogates, and physics-discovery methods require different data structures, labels, and validation protocols. A useful benchmark ecosystem should therefore define several linked tasks rather than one universal leaderboard.

\begin{landscape}
\begingroup
\scriptsize
\setlength{\tabcolsep}{3pt}
\renewcommand{\arraystretch}{1.18}
\begin{longtable}{>{\raggedright\arraybackslash}p{0.18\linewidth} >{\raggedright\arraybackslash}p{0.11\linewidth} >{\raggedright\arraybackslash}p{0.24\linewidth} >{\raggedright\arraybackslash}p{0.20\linewidth} >{\raggedright\arraybackslash}p{0.20\linewidth}}
\caption{Review synthesis matrix for two-phase heat-transfer AI benchmarks.}\label{tab:4}\\
\toprule
AI/data direction & Typical S+TD class & Representative inputs and labels & Benchmark tasks & Metadata and split requirements \\
\midrule
\endfirsthead
\caption[]{Review synthesis matrix for two-phase heat-transfer AI benchmarks. (continued)}\\
\toprule
AI/data direction & Typical S+TD class & Representative inputs and labels & Benchmark tasks & Metadata and split requirements \\
\midrule
\endhead
\midrule\multicolumn{5}{r}{Continued on next page}\\
\endfoot
\bottomrule
\endlastfoot
Tabular HTC, pressure-drop, flow-regime, and CHF prediction & 0+0D, 0+1D & Operating conditions, fluid properties, geometry, surface descriptors, scalar performance labels & HTC regression, CHF-margin prediction, pressure-drop regression, regime classification & Fluid-held-out, geometry-held-out, facility-held-out, and correlation-baseline splits; uncertainty and definition of each scalar label \\
Optical boiling and two-phase-flow vision & 2+0D, 2+1D & High-speed images/videos, bubble masks, flow-pattern labels, bubble statistics & Bubble segmentation, tracking, regime classification, bubble-frequency prediction, optical-to-performance inference & Pixel scale, frame rate, illumination, view angle, surface condition, annotation protocol, video-level rather than frame-level splits \\
Infrared thermometry and thermal-field learning & 2+0D, 2+1D & IR frames, wall-temperature fields, inferred heat flux, dry-area masks, CHF precursors & Dryout detection, heat-flux reconstruction, boiling-crisis warning, temperature-to-mechanism mapping, image-to-thermal-field inference & Emissivity, calibration, substrate/heater model, inverse-method details, uncertainty, synchronized optical/acoustic records where available \\
Acoustic and acoustic-emission diagnostics & 0+1D, multimodal S+TD & Hydrophone/AE waveforms, hit features, spectrograms, heat flux, regime, CHF warning labels & Acoustic heat-flux regression, regime classification, early-warning detection, visual-limited monitoring & Sensor placement, bandwidth, coupling, sampling rate, synchronization, background noise, run-level splits to prevent leakage \\
Simulation, CFD, and scientific ML & 2+1D, 3+0D, 3+1D & Interface-resolved fields, pressure/velocity/temperature fields, phase indicators, reduced-order states & Surrogate modeling, neural operators, closure learning, experiment-simulation transfer & Boundary conditions, mesh/time-step information, solver version, closure assumptions, parameter sweeps, out-of-distribution test cases \\
Inverse modeling and governing-physics discovery & 0+1D, 1+1D, 2+1D & Raw temperature histories, sensor geometry, reconstructed heat flux, candidate state variables & Inverse heat-conduction recovery, equation discovery, uncertainty-aware model identification & Raw measurements before reduction, sensor coordinates, priors/regularization, repeated conditions, noise models, physical residual checks \\
Multimodal fusion and foundation-style pretraining & mixed S+TD & Synchronized optical, IR, acoustic, AE, scalar sensors, CFD, and metadata & Missing-modality robustness, cross-modal prediction, self-supervised representation learning, transfer between facilities & Synchronization metadata, common event clocks, modality dropout tests, lab-held-out and surface-held-out evaluation \\
\end{longtable}
\endgroup
\end{landscape}

This matrix also clarifies why benchmark datasets must preserve raw data and processing provenance. For example, an IR benchmark that provides only final heat-flux maps cannot evaluate alternative inverse heat-conduction models; an acoustic benchmark that provides only spectrograms cannot test whether waveform-level learning improves early warning; and an image benchmark split frame-by-frame can overestimate generalization because neighboring frames from the same video are strongly correlated. The benchmark unit should usually be a run, surface, fluid, geometry, or laboratory, not an individual frame or row.

A practical roadmap for two-phase heat-transfer AI can progress through five stages. The first stage is dataset rescue and release. Many valuable datasets already exist in laboratories as camera files, acoustic recordings, infrared images, spreadsheets, CFD outputs, and instrument-specific formats. The near-term priority is to release these records with persistent identifiers, minimal but complete metadata, license information, raw-to-processed data provenance, and executable decoder or preprocessing scripts. At this stage, the goal is not to create perfect benchmarks; it is to prevent high-cost experimental and simulation data from remaining invisible or unusable.

The second stage is physics-aware standardization. Released datasets should be converted into self-describing formats and accompanied by metadata that records not only file type, date, and authorship, but also physical context, including working fluid, pressure, mass flux, heat flux, surface material, roughness, wettability, heater dimensions, channel geometry, subcooling, vapor quality, sensor locations, calibration, uncertainty, and synchronization. Dimensionless groups such as Reynolds number, Weber number, Boiling number, Jakob number, Capillary number, Bond number, confinement number, and CHF-normalized heat flux should be computed where appropriate. This stage transforms a collection of files into a searchable physical dataset.

The third stage is benchmark construction. Benchmark datasets should define canonical tasks, such as boiling-regime classification, bubble segmentation, heat-flux regression, CHF-margin prediction, HTC prediction, pressure-drop prediction, optical-to-thermal inference, acoustic-to-regime classification, and CFD surrogate modeling. Each benchmark should include baseline correlations, baseline ML models, predefined train-validation-test splits, uncertainty metrics, and failure-case examples. Splits should test physically meaningful generalization, including surface-held-out, fluid-held-out, geometry-held-out, heat-load-path-held-out, and laboratory-held-out evaluations. This stage is where datasets become useful for comparing methods rather than only demonstrating individual models.

The fourth stage is multimodal and multi-fidelity learning. Once individual benchmarks are stable, the field can build larger datasets that align optical videos, IR fields, acoustic/AE signals, pressure and temperature histories, heat-flux labels, CFD fields, and reduced-order simulation outputs. These datasets can support sensor fusion, self-supervised pretraining, transfer learning, uncertainty-aware inference, and hybrid experiment-simulation learning. The most valuable models at this stage will not simply reduce mean prediction error; they will learn representations that remain physically meaningful when sensors are missing, operating conditions shift, or experimental facilities differ.

The fifth stage is community governance and living infrastructure. Mature two-phase AI should have curated databanks, versioned benchmark releases, model cards, dataset datasheets, leaderboards with leakage controls, citation standards, and review processes for contributed data and software. Community hubs such as Multiphase Hub, TFTL's dataset page, BubbleML, NED3 datasets, and Bubble Dynamics Research Tools are early steps in this direction. The long-term objective is a living ecosystem in which new experiments and simulations can be added without breaking benchmark comparability, and in which models are evaluated not only by accuracy, but also by reproducibility, physical consistency, uncertainty, interpretability, and transferability.

\begin{figure}[H]
\centering
\includegraphics[width=0.95\linewidth]{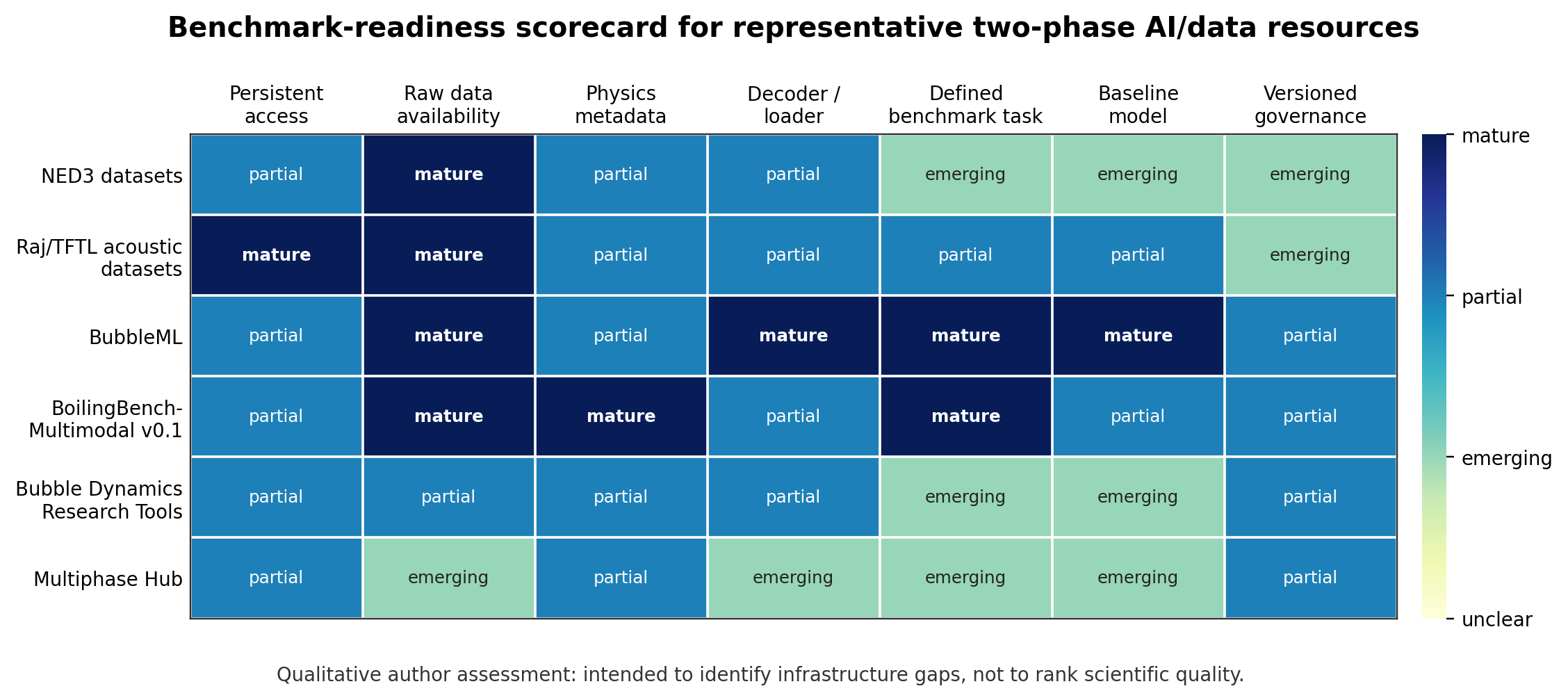}
\caption{Benchmark-readiness scorecard for representative two-phase AI and data resources. The chart evaluates persistent access, raw-data availability, physics metadata, decoders/loaders, defined benchmark tasks, baseline models, and versioned governance using qualitative categories: unclear, emerging, partial, and mature. These categories indicate whether each infrastructure component is publicly verifiable, independently reusable, and governed through stable versions or documentation. The assessment is intended to expose infrastructure gaps and motivate community coordination, not to rank the underlying scientific contributions.}
\label{fig:readiness_scorecard}
\end{figure}

Figure~\ref{fig:readiness_scorecard} translates these requirements into a practical readiness scorecard for representative two-phase AI/data resources. The scorecard is intentionally qualitative and should not be read as a ranking of scientific merit. The categories are assigned using the following rubric. ``Unclear'' means that the component could not be verified from public documentation. ``Emerging'' means that the component is mentioned or partially visible but is not yet sufficient for independent reuse. ``Partial'' means that the component is available for some tasks, subsets, or examples, but lacks full coverage, standardization, or versioned governance. ``Mature'' means that the component is openly available, documented, versioned or citable, and usable by an independent researcher without private communication. For example, an open dataset may have strong raw-data availability but only partial benchmark governance, while a hub may provide useful discovery and curation but not yet define fixed train-test splits or baseline models. A mature ecosystem will need both archives that preserve expensive data and benchmark layers that make model comparison fair.

\section{Future Work and Community Effort}

\subsection{Benchmark dataset development}

Open datasets become most useful when they are paired with benchmark tasks that are clear enough to reproduce and flexible enough to invite better methods. The immediate community need is not one definitive leaderboard, but several seed benchmarks with explicit task definitions, data modalities, input-output mappings, train-validation-test splits, evaluation metrics, and physical interpretation. BoilingBench-Multimodal v0.1 is discussed here only as an initial example of this mechanism because it defines run-level splits, metadata, manifests, leakage rules, contribution guidelines, and Zenodo-oriented data staging for multimodal boiling datasets [89]. It should not be read as a completed community standard. Its role is to demonstrate the components that a reproducible benchmark needs and to provide a starting point for community audit, extension, and replacement.

A useful benchmark program should test physically meaningful generalization. Generalization across random windows from one run is different from generalization across heat-flux ramps, surfaces, fluids, pressures, geometries, and laboratories. For this reason, seed benchmarks should report run-held-out, surface-held-out, fluid-held-out, geometry-held-out, heat-load-path-held-out, and laboratory-held-out splits when the available data permit them. Each benchmark should also include raw files, processed files, metadata tables, synchronization records, baseline notebooks, reference metrics, and documented failure cases.

Reviewer comments on this manuscript also expose an important next step for the field: quantitative comparison across S+TD modalities. A fair comparison should not simply train unrelated models on unrelated datasets and compare headline accuracy. It should define a common prediction target, comparable training and test splits, uncertainty-aware metrics, and computational-cost reporting. For example, a multimodal boiling benchmark could compare hydrophone-to-heat-flux regression, AE-hit-to-heat-flux regression, image-sequence-to-heat-flux regression, and multimodal fusion on matched runs. Reported metrics should include MAE, RMSE, relative error, error near transition events, inference cost, training cost, and performance under held-out surfaces or heat-load paths. The present review therefore treats cross-modality experiments as a benchmark objective that the community should complete using openly released datasets, rather than as a new experiment performed within this narrative review.

The proposed benchmark program should be staged. Early releases can rescue and document valuable datasets; later releases can add standardized loaders, baseline models, versioned splits, and governance. Future versions could add datasets from TFTL, BubbleML, Multiphase Hub, synchronized optical/IR boiling measurements, infrared boiling datasets, optical and event-vision datasets, and other community contributions. A lightweight governance model could include public issue tracking, required metadata templates, contributor checklists, versioned releases, dataset and software DOIs, baseline-model pull requests, and periodic benchmark reports. In this model, the present paper proposes one possible starting mechanism; the community must complete, audit, and expand it.

\subsection{Community databanks and one-stop AI/ML tool libraries}

The long-term goal is larger than releasing individual datasets. Thermal-fluid AI needs shared databanks that allow independent groups to contribute, discover, compare, and reuse data across mechanisms, diagnostics, and applications. A databank differs from a static data repository because it provides curation, metadata conventions, dataset indexing, benchmark tasks, tool links, versioning, and community contribution pathways. Multiphase Hub provides an example of this direction for multiphase transport. It describes itself as a centralized resource for accessing, sharing, and analyzing multiphase transport data, with datasets, analysis, flow-pattern data, physical-property measurements, modeling tools, and visualization tools [90]. For multiphase transport and thermal management, future databanks could organize datasets by fluid, geometry, surface, pressure, heat-flux range, Reynolds number, Boiling number, Weber number, instrumentation, S+TD class, and target ML task.

Community databanks would also reduce duplication. Many laboratories collect time series, images, videos, IR fields, acoustic signals, and CFD fields, but each group often builds its own file conventions and analysis scripts. A shared framework would allow users to ask practical questions: Which datasets contain synchronized hydrophone and heat-flux measurements? Which include high-speed boiling videos on copper microchannels? Which provide annotated images for bubble segmentation? Which contain 3D CFD fields suitable for surrogate modeling? Which datasets are appropriate for cross-laboratory generalization tests? These are discoverability problems as much as storage problems.

\begin{figure}[H]
\centering
\includegraphics[width=0.95\linewidth]{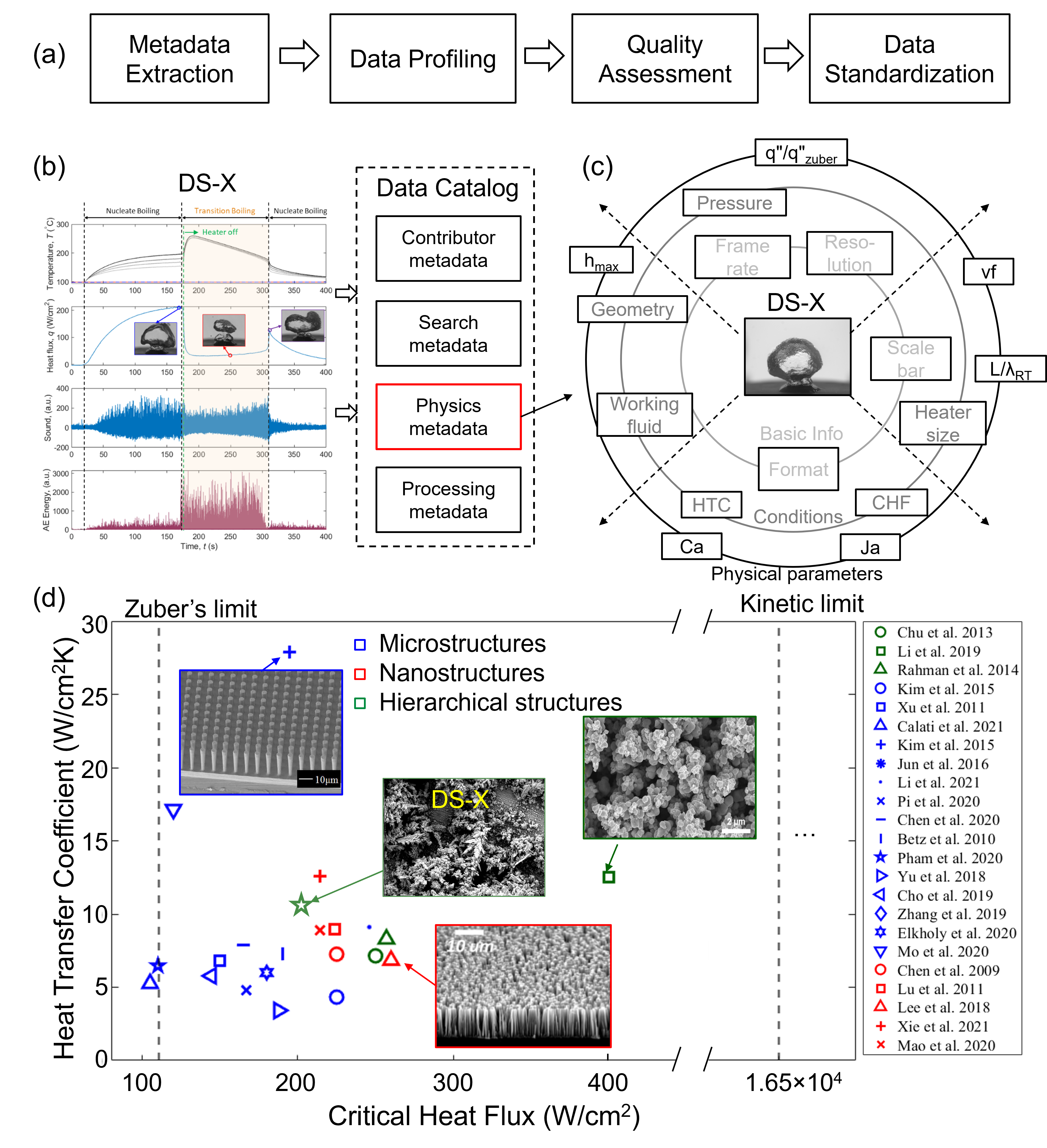}
\caption{Proposed data curation and physics metadata architecture for a two-phase cooling dataset. (a) Workflow from metadata extraction to data profiling, quality assessment, and standardization. (b) Example DS-X upload in which contributor metadata, search metadata, physics metadata, and processing metadata are extracted and added to the data catalog. (c) Physics-metadata representation of a boiling video using basic information, operating conditions, performance metrics, and mechanism-level dimensionless parameters. (d) Two-dimensional projection of the high-dimensional dataset profile in heat-transfer-coefficient versus critical-heat-flux space, where DS-X is compared with literature HTC/CHF data and physical-limit references [196-217].}
\label{fig:curation_metadata}
\end{figure}

The architecture in Figure~\ref{fig:curation_metadata} should be read as a proposed data-curation architecture rather than an existing community standard. Elements that already exist include public dataset pages, archival records, selected open-source decoders, and community resources such as Multiphase Hub. Elements under development include benchmark manifests, metadata templates, and seed benchmark governance. Elements proposed for future community work include automated metadata scrapers, high-dimensional dataset profiling, physics-law consistency checks, standardized MTDL-style protocols, and warehouse-scale contribution loops.

A community databank should include a data-curation layer that is more active than a normal catalog. Figure~\ref{fig:curation_metadata}a illustrates the proposed curation workflow using a two-phase cooling dataset as an example. After a contributor uploads a new dataset, denoted DS-X, through the user interface, the system first performs metadata extraction to create dataset labels and add the dataset to the catalog. The extracted metadata are then used to profile DS-X in a high-dimensional parameter space together with existing datasets already in the system. Dataset quality is assessed by examining two forms of consistency: consistency with related datasets in the data catalog and consistency with fundamental physical laws. Finally, DS-X is standardized by converting raw and processed data into formats prescribed by Multiphase Thermal-fluid Data Library (MTDL) protocols and storing the standardized files in the data warehouse.

The distinguishing feature of this system is physics metadata. A normal catalog records general information such as title, authors, file type, date, keywords, and license. The proposed MTDL catalog should additionally log the physical and experimental context that determines whether a dataset is reusable for thermal-fluid AI. For a high-speed boiling video in DS-X, physics metadata would include basic acquisition information such as resolution, scale bar, frame rate, and file format; operating and performance conditions such as working fluid, heater size, geometry, pressure, heat-transfer coefficient, and critical heat flux; and mechanism-level parameters such as the ratio of heater size to the critical Rayleigh-Taylor wavelength, Jakob number, Capillary number, vapor fraction, maximum heat-transfer coefficient, and Zuber-limit-scaled heat flux (Figure~\ref{fig:curation_metadata}b). These fields convert a video from an isolated media file into a searchable physical record.

Automated metadata extraction should combine contributor-entered metadata with algorithmic scrapers. Simple scrapers can mine headings, column names, units, time stamps, spatial coordinates, geographic tags, file formats, and acquisition settings from spreadsheets, text files, images, and instrument exports. Domain-aware scrapers can match extracted terms against an ontology for boiling, two-phase flow, thermal management, acoustics, infrared thermography, CFD, or other relevant disciplines. More advanced feature-detection algorithms can compute data-derived metadata, including heat-flux ranges from time histories, bubble-size distributions from videos, spectral features from acoustic signals, temperature-field extrema from infrared images, and operating regimes from synchronized multimodal records. The same idea generalizes beyond thermal-fluid data. Rainfall minima, maxima, and averages can be extracted from environmental datasets; spatial and temporal features can be mined from satellite images; and discipline-specific workflows can extract storms, astronomical objects, or social-event patterns. In the proposed curation system, the Data Curator should allow new feature-detection algorithms to be registered, associated with dataset types, and automatically run to add contextual features into the relevance-network knowledge base.

Once curated, DS-X can be mapped into the same high-dimensional parameter space as existing datasets (Figure~\ref{fig:curation_metadata}c). This profile enables semantic search, similarity matching, outlier detection, and benchmark selection. For example, a user could query for boiling datasets with similar pressure, working fluid, heater scale, CHF range, optical frame rate, or acoustic bandwidth. Thus, the system could flag DS-X if its reported heat flux, surface geometry, or dimensionless groups appear inconsistent with neighboring datasets or known physical limits. Figure~\ref{fig:curation_metadata}d shows a representative two-dimensional projection of this high-dimensional profile in heat-transfer-coefficient versus critical-heat-flux space. In that projection, DS-X is marked with a green star and compared with literature data from studies of nanowire, mixed-wettability, micropillar, microporous, metal-foam, 3D-printed, graphene-oxide, and hierarchical boiling surfaces [196-217]. Such projections do not replace the full metadata profile, but they give contributors and users an interpretable view of where a new dataset sits relative to known boiling-performance regimes, including Zuber-type and kinetic-limit references.

One-stop tool libraries are the software counterpart to databanks. They can help users discover compatible decoders, visualization scripts, segmentation models, acoustic-analysis tools, sequence-regression workflows, CFD surrogate-modeling utilities, and benchmark baselines. Future libraries should distinguish peer-reviewed software, maintained packages, educational examples, paper artifacts, datasets, and emerging candidate repositories, while recording scope, input/output data types, validation status, license, maintenance activity, installation difficulty, and known limitations.

The Bubble Dynamics Research Tools repository provides a bubble-focused example of this hub model [91]. It curates datasets, code, and analysis tools for bubble dynamics research, including vapor bubbles during boiling, gas bubbles in liquids, optical imaging, acoustic emissions, thermal imaging, CFD, and scientific machine-learning simulations. It complements the Data Center Cooling Research Tools collection [88] by narrowing the scope to bubble datasets and bubble-dynamics analysis packages, including resources for segmentation, tracking, acoustic analysis, thermal diagnostics, reduced-order bubble dynamics, interface-resolved simulation, and ML benchmarking. Such domain-specific hubs can lower the entry barrier for new users while preserving a screening layer that distinguishes reusable tools and datasets from preliminary code artifacts.

\subsection{Limitations and open challenges}

Open thermal-fluid datasets face limitations that should be stated clearly. First, experimental datasets are often specific to a facility, geometry, sensor arrangement, and operating protocol. Models trained on one dataset may not generalize to new surfaces, fluids, pressures, heaters, flow channels, or camera views. Second, multimodal synchronization is difficult. Small timing offsets can matter when transient events are fast relative to sensor response or frame rate. Third, labels such as heat flux, regime, or event timing are often derived from models or thresholds, not directly measured ground truth. Fourth, data volume can be large, especially for videos, infrared sequences, waveform streams, and 3D CFD fields. Fifth, some proprietary instrument formats may require reverse engineering or vendor cooperation before fully open decoding is possible.

These limitations do not weaken the case for open datasets; they define the work that open infrastructure makes possible. When datasets are public, the community can test robustness, compare preprocessing methods, identify leakage, quantify uncertainty, and improve models. When software is public, users can inspect assumptions, adapt tools to new instruments, and contribute fixes. When benchmarks are public, claims about model performance become easier to evaluate and reproduce.

\section{Conclusions}

This review presents open data and machine learning for two-phase heat transfer as an infrastructure problem as much as a modeling problem. The central organizing idea is the S+TD taxonomy, which classifies data by spatial and temporal dimensionality and connects thermal-fluid measurements to appropriate AI/ML workflows. The review shows that two-phase AI spans tabular correlation learning, optical imaging, infrared thermometry, acoustic and acoustic-emission sensing, event streams, CFD-derived fields, scientific machine learning, and multimodal fusion.

The reviewed resources show that paired dataset and software release is becoming a practical pattern rather than a single-laboratory model. Public acoustic, optical, infrared, simulation, and multimodal datasets are now accompanied by tools for segmentation, sequence regression, waveform decoding, thermographic analysis, surrogate modeling, and benchmark construction. Emerging hubs and repositories, including Multiphase Hub, BubbleML, Bubble Dynamics Research Tools, NED3 dataset releases, and related laboratory or community collections, show that the field is moving toward a distributed open-data ecosystem.

The proposed community direction is to make two-phase heat-transfer data findable, decodable, benchmarkable, and physically interpretable. Achieving this goal will require shared physics metadata, versioned dataset releases, open processing tools, baseline examples, careful benchmark splits, uncertainty reporting, and curated tool hubs. The strongest future datasets will not only store files; they will preserve the physical context needed to evaluate whether an AI model has learned transferable transport behavior.

\section*{Data and software availability}

The NED3 open-source software packages are summarized at \url{https://ned3.uark.edu/software/} [72]. The NED3 open datasets are summarized at \url{https://ned3.uark.edu/datasets/} [73]. SeqReg is available at \url{https://github.com/cldunlap73/SeqReg} [87]. BubbleID is available at \url{https://github.com/cldunlap73/BubbleID} [29]. CFDTwin is available at \url{https://github.com/UARK-NED3/CFDTwin} [72]. Bubbleformer is available at \url{https://github.com/HPCForge/bubbleformer/tree/main} [41]. BubbleML 2.0 is available through Zenodo and Hugging Face [42-43]. NUCLEUS is available at \url{https://github.com/therml-ai/NUCLEUS} [50]. The HB-ARFM repository is available at \url{https://github.com/therml-ai/HB-ARFM} [47]. The optical-to-infrared condensation mapping code and sample data are available at \url{https://github.com/SiaK4/cGAN-image-to-temperature-mapping} [86]. The NED3 Data Center Cooling Research Tools collection is available at \url{https://github.com/UARK-NED3/Data-Center-Cooling-Research-Tools} [88]. The NED3 Bubble Dynamics Research Tools collection is available at \url{https://github.com/UARK-NED3/Bubble-Dynamics-Research-Tools} [91]. BoilingBench-Multimodal v0.1 is available at \url{https://github.com/UARK-NED3/BoilingBench-Multimodal} [89], with the full raw-data archive intended for Zenodo deposition.

\section*{Acknowledgments}

This study was supported by the National Science Foundation, United States, under Grant No. CBET-2323022, TI-2431969, and OIA-1946391.

\section*{References}
\begin{enumerate}[label={[\arabic*]}, leftmargin=*]

\item W. M. Rohsenow, "A method of correlating heat-transfer data for surface boiling of liquids," Transactions of the ASME, vol. 74, no. 6, pp. 969-975, 1952. (\url{https://doi.org/10.1115/1.4015984})

\item S. S. Kutateladze, Heat Transfer in Condensation and Boiling, AEC-tr-3770, United States Atomic Energy Commission, 1952. ISBN not assigned in the AEC translation record.

\item N. Zuber, "Hydrodynamic aspects of boiling heat transfer," Ph.D. dissertation, University of California, Los Angeles, 1959. (\url{https://doi.org/10.2172/4175511})

\item V. K. Dhir, "Boiling heat transfer," Annual Review of Fluid Mechanics, vol. 30, pp. 365-401, 1998. (\url{https://doi.org/10.1146/annurev.fluid.30.1.365})

\item V. P. Carey, Liquid-Vapor Phase-Change Phenomena, 2nd ed., Taylor \& Francis, 2008 (ISBN: 978-1-59169-035-1).

\item J. Kim, "Review of nucleate pool boiling bubble heat transfer mechanisms," International Journal of Multiphase Flow, vol. 35, no. 12, pp. 1067-1076, 2009. (\url{https://doi.org/10.1016/j.ijmultiphaseflow.2009.07.008})

\item M. G. Cooper, "Heat flow rates in saturated nucleate pool boiling: A wide-ranging examination using reduced properties," Advances in Heat Transfer, vol. 16, pp. 157-239, 1984.

\item M. Ravichandran and M. Bucci, "Online, quasi-real-time analysis of high-resolution, infrared, boiling heat transfer investigations using artificial neural networks," Applied Thermal Engineering, vol. 163, 114357, 2019. (\url{https://doi.org/10.1016/j.applthermaleng.2019.114357})

\item M. Ravichandran, G. Su, C. Wang, J. H. Seong, A. Kossolapov, B. Phillips, M. M. Rahman, and M. Bucci, "Decrypting the boiling crisis through data-driven exploration of high-resolution infrared thermometry measurements," Applied Physics Letters, vol. 118, 253903, 2021. (\url{https://doi.org/10.1063/5.0048391})

\item MIT News, "Infrared cameras and artificial intelligence provide insight into boiling," July 7, 2021. \url{https://news.mit.edu/2021/infrared-cameras-artificial-intelligence-provide-insight-into-boiling-0707}

\item L. Zhang, C. Wang, G. Su, A. Kossolapov, G. M. Aguiar, J. H. Seong, F. Chavagnat, B. Phillips, M. M. Rahman, and M. Bucci, "A unifying criterion of the boiling crisis," Nature Communications, vol. 14, 2321, 2023. (\url{https://doi.org/10.1038/s41467-023-37899-7})

\item M. Ravichandran, A. Kossolapov, G. M. Aguiar, B. Phillips, and M. Bucci, "Autonomous and online detection of dry areas on a boiling surface using deep learning and infrared thermometry," Experimental Thermal and Fluid Science, vol. 145, 110879, 2023. (\url{https://doi.org/10.1016/j.expthermflusci.2023.110879})

\item G. M. Hobold and A. K. da Silva, "Machine learning classification of boiling regimes with low speed, direct and indirect visualization," International Journal of Heat and Mass Transfer, vol. 125, pp. 1296-1309, 2018. (\url{https://doi.org/10.1016/j.ijheatmasstransfer.2018.04.156})

\item R. Raj, "Dataset: Deep learning the sound of boiling for advance prediction of boiling crisis," Mendeley Data, Version 1, 2021. (\url{https://doi.org/10.17632/vnnzc7n97m.1})

\item K. N. R. Sinha, V. Kumar, N. Kumar, A. Thakur, and R. Raj, "Dataset for boiling acoustic emissions: A tool for data driven boiling regime prediction," Data in Brief, vol. 52, 109793, 2024. (\url{https://doi.org/10.1016/j.dib.2023.109793})

\item B. Suriyaprasaad, A. Upadhyay, A. Thakur, and R. Raj, "A roadmap for decoding the sound of boiling," npj Thermal Science and Engineering, vol. 1, 2, 2026. (\url{https://doi.org/10.1038/s44435-025-00004-z})

\item C. Dunlap, H. Pandey, E. Weems, and H. Hu, "Data for: Nonintrusive heat flux quantification using acoustic emissions during pool boiling," Dryad, 2025. (\url{https://doi.org/10.5061/dryad.q573n5tvq})

\item C. Dunlap, C. Li, H. Pandey, and H. Hu, "Data for: Hit2Flux: A machine learning framework for boiling heat flux prediction using hit-based acoustic emission sensing," Dryad, 2025. (\url{https://doi.org/10.5061/dryad.g79cnp628})

\item S. M. S. Hassan, A. Feeney, A. Dhruv, J. Kim, Y. Suh, J. Ryu, Y. Won, and A. Chandramowlishwaran, "BubbleML: A Multi-Physics Dataset and Benchmarks for Machine Learning," arXiv:2307.14623, 2023. (\url{https://doi.org/10.48550/arXiv.2307.14623})

\item D. Garg, A. Bard, Y. Qiu, C. R. Kharangate, and R. H. French, "Machine learning algorithms to predict flow boiling pressure drop in mini/micro-channels based on universal consolidated data," International Journal of Heat and Mass Transfer, vol. 178, 121607, 2021. (\url{https://doi.org/10.1016/j.ijheatmasstransfer.2021.121607})

\item Q. Fu, Y. Suh, X. Zhang, S. Chang, and Y. Won, "Bubble2Heat: Optical to Thermal Inference in Pool Boiling Using Physics-encoded Generative AI," arXiv:2505.00823, 2025. \url{https://arxiv.org/abs/2505.00823}

\item S. Chang, S. Arani, N. S. Nuthalapati, Y. Suh, N. Choi, S. Khodakarami, M. R. H. Roni, N. Miljkovic, A. Chandramowlishwaran, and Y. Won, "EventFlow: Real-time neuromorphic event-driven classification of two-phase boiling flow regimes," Droplet, 2026. (\url{https://doi.org/10.1002/dro2.70066})

\item K. N. R. Sinha, V. Kumar, N. Kumar, A. Thakur, and R. Raj, "Deep learning the sound of boiling for advance prediction of boiling crisis," Cell Reports Physical Science, vol. 2, 100382, 2021. (\url{https://doi.org/10.1016/j.xcrp.2021.100382})

\item C. Dunlap, C. Li, H. Pandey, Y. Sun, and H. Hu, "Data for: A temporal-spatial framework for efficient heat flux monitoring of transient boiling," Dryad, 2025. (\url{https://doi.org/10.5061/dryad.6m905qgc7})

\item H. Pandey, C. Li, C. Dunlap, and H. Hu, "Data from: Unveiling hysteresis of transient boiling: A multimodal perspective," Dryad, 2025. (\url{https://doi.org/10.5061/dryad.ksn02v7h2})

\item C. Dunlap, H. Pandey, E. Weems, and H. Hu, "Nonintrusive Heat Flux Quantification Using Acoustic Emissions During Pool Boiling," Applied Thermal Engineering, 228, 120558, 2023. (\url{https://doi.org/10.1016/j.applthermaleng.2023.120558})

\item M. Bongarala, J. A. Weibel, and S. V. Garimella, "Boiling crisis is a consequence of thermal runaway in the substrate due to degradation of nucleate boiling heat transfer," International Journal of Heat and Mass Transfer, vol. 233, 125906, 2024. (\url{https://doi.org/10.1016/j.ijheatmasstransfer.2024.125906})

\item A. Rokoni, L. Zhang, T. Soori, H. Hu, T. Wu, and Y. Sun, "Learning new physical descriptors from reduced-order analysis of bubble dynamics in boiling heat transfer," International Journal of Heat and Mass Transfer, vol. 186, 122501, 2022. (\url{https://doi.org/10.1016/j.ijheatmasstransfer.2021.122501})

\item C. Dunlap and contributors, "BubbleID," GitHub repository. \url{https://github.com/cldunlap73/BubbleID}

\item C. Dunlap, C. Li, H. Pandey, N. Le, and H. Hu, "Data from: BubbleID: A deep learning framework for bubble interface dynamics analysis," Dryad, 2025. (\url{https://doi.org/10.5061/dryad.ksn02v7gx})

\item Y. Gao, H.-X. Yu, B. Zhu, and J. Wu, "FluidNexus: 3D Fluid Reconstruction and Prediction from a Single Video," Proceedings of the IEEE/CVF Conference on Computer Vision and Pattern Recognition, 2025. \url{https://github.com/ueoo/FluidNexus}

\item Y. Suh, S. Chang, P. Simadiris, T. B. Inouye, M. J. Hoque, S. Khodakarami, C. R. Kharangate, N. Miljkovic, and Y. Won, "VISION-iT: A Framework for Digitizing Bubbles and Droplets," Energy and AI, vol. 15, 100309, 2024. (\url{https://doi.org/10.1016/j.egyai.2023.100309})

\item V. Serdyukov, R. Mullyadzanov, and A. Surtaev, "Deep learning segmentation to analyze bubble dynamics and heat transfer during boiling at various pressures," International Journal of Multiphase Flow, 104402, 2023. (\url{https://doi.org/10.1016/j.ijmultiphaseflow.2023.104402})

\item R. Mosdorf and G. Gorski, "Identification of two-phase flow patterns in minichannel based on RQA and PCA analysis," International Journal of Heat and Mass Transfer, vol. 96, pp. 64-74, 2016. (\url{https://doi.org/10.1016/j.ijheatmasstransfer.2016.01.015})

\item HPCForge, "BubbleML: A multiphase multiphysics dataset and benchmarks for scientific machine learning," GitHub repository. \url{https://github.com/HPCForge/BubbleML}

\item H. Zhai, Q. Zhou, and G. Hu, "Predicting micro-bubble dynamics with semi-physics-informed deep learning," AIP Advances, vol. 12, 035153, 2022. (\url{https://doi.org/10.1063/5.0079602})

\item comp-physics, "bubble-dynamics-resnet: Integrate bubble dynamics faster," GitHub repository. \url{https://github.com/comp-physics/bubble-dynamics-resnet}

\item BerryWei, "Bubble-dynamic: Rayleigh Plesset \& Keller Miksis equation," GitHub repository. \url{https://github.com/BerryWei/Bubble-dynamic}

\item kozakaron, "Bubble\_dynamics\_simulation," GitHub repository. \url{https://github.com/kozakaron/Bubble_dynamics_simulation}

\item S. M. S. Hassan, X. Zou, A. Dhruv, V. Ganesan, and A. Chandramowlishwaran, "Bubbleformer: Forecasting boiling with transformers," Advances in Neural Information Processing Systems, Datasets and Benchmarks Track, 2025. \url{https://papers.neurips.cc/paper_files/paper/2025/file/18d2e6c115da385ff6a7ac53655ae95e-Paper-Datasets_and_Benchmarks_Track.pdf}

\item HPCForge, "Bubbleformer: Forecasting Boiling with Transformers," GitHub repository. \url{https://github.com/HPCForge/bubbleformer/tree/main}

\item A. Chandramowlishwaran, A. Dhruv, X. Zou, and V. Ganesan, "Bubbleformer: Forecasting Boiling with Transformers," Zenodo dataset, Version 0.0, 2025. (\url{https://doi.org/10.5281/zenodo.15429313})

\item HPCForge, "BubbleML\_2," Hugging Face dataset. \url{https://huggingface.co/datasets/hpcforge/BubbleML_2}

\item D. Sculley et al., "Hidden technical debt in machine learning systems," Advances in Neural Information Processing Systems, vol. 28, 2015.

\item O. E. Gundersen and S. Kjensmo, "State of the art: Reproducibility in artificial intelligence," Proceedings of the AAAI Conference on Artificial Intelligence, vol. 32, no. 1, 2018. (\url{https://doi.org/10.1609/aaai.v32i1.11503})

\item X. Zou, S. M. S. Hassan, A. Feeney, and A. Chandramowlishwaran, "History-Bootstrapped Flow Matching for Inverse Boiling Reconstruction," arXiv:2606.00349, 2026. (\url{https://doi.org/10.48550/arXiv.2606.00349})

\item therml-ai, "HB-ARFM: History-Bootstrapped Flow Matching for Inverse Boiling Reconstruction," GitHub repository. \url{https://github.com/therml-ai/HB-ARFM}

\item H. Wang, R. Li, F. Xu, F. Sun, K. Han, Z. Huang, C. Chang, X. Luo, W. Wang, and Y. Sun, "FD-Bench: A Modular and Fair Benchmark for Data-driven Fluid Simulation," arXiv:2505.20349, 2025. (\url{https://doi.org/10.48550/arXiv.2505.20349})

\item A. Feeney, X. Zou, S. M. S. Hassan, S. Rachabathuni, and A. Chandramowlishwaran, "NUCLEUS-MoE: Unified Model of Pool Boiling for Liquid Cooling," arXiv:2605.27722, 2026. (\url{https://doi.org/10.48550/arXiv.2605.27722})

\item therml-ai, "NUCLEUS: ML Surrogates for pool boiling," GitHub repository. \url{https://github.com/therml-ai/NUCLEUS}

\item J. E. Taylor, "bubblemask: Python package for applying Gaussian Bubbles masks to image stimuli," GitHub repository. \url{https://github.com/JackEdTaylor/bubblemask}

\item C. Dunlap, C. Li, H. Pandey, Y. Sun, and H. Hu, "A Temporal-Spatial Framework for Efficient Heat Flux Monitoring of Transient Boiling," IEEE Transactions on Instrumentation and Measurement, 2024. (\url{https://doi.org/10.1109/TIM.2024.3460944})

\item C. Dunlap, C. Li, H. Pandey, and H. Hu, "Hit2Flux: A Machine Learning Framework for Boiling Heat Flux Prediction Using Hit-Based Acoustic Emission Sensing," AI Thermal Fluids, 1, 100002, 2025. (\url{https://doi.org/10.1016/j.aitf.2025.100002})

\item H. Pandey, C. Li, C. Dunlap, and H. Hu, "Unveiling Hysteresis of Transient Boiling: A Multimodal Perspective," Applied Thermal Engineering, 262, 125259, 2025.

\item M. H. Mousa, C.-M. Yang, K. Nawaz, and N. Miljkovic, "Review of heat transfer enhancement techniques in two-phase flows for highly efficient and sustainable cooling," Renewable and Sustainable Energy Reviews, vol. 155, 111896, 2022. (\url{https://doi.org/10.1016/j.rser.2021.111896})

\item Y. Suh, J. Lee, P. Simadiris, X. Yan, S. Sett, L. Li, K. Fazle Rabbi, N. Miljkovic, and Y. Won, "A deep learning perspective on dropwise condensation," Advanced Science, vol. 8, no. 22, 2101794, 2021. (\url{https://doi.org/10.1002/advs.202101794})

\item M. D. Wilkinson et al., "The FAIR Guiding Principles for scientific data management and stewardship," Scientific Data, vol. 3, 160018, 2016. (\url{https://doi.org/10.1038/sdata.2016.18})

\item M. Altman and G. King, "A proposed standard for the scholarly citation of quantitative data," D-Lib Magazine, vol. 13, no. 3/4, 2007. (\url{https://doi.org/10.1045/march2007-altman})

\item M. Crosas, "The Dataverse Network: An open-source application for sharing, discovering and preserving data," D-Lib Magazine, vol. 17, no. 1/2, 2011. (\url{https://doi.org/10.1045/january2011-crosas})

\item C. Mayo, T. J. Vision, and E. A. Hull, "The location of the citation: Changing practices in how publications cite original data in the Dryad Digital Repository," International Journal of Digital Curation, vol. 11, no. 1, pp. 150-155, 2016. (\url{https://doi.org/10.2218/ijdc.v11i1.400})

\item E. D. Foster and A. Deardorff, "Open Science Framework (OSF)," Journal of the Medical Library Association, vol. 105, no. 2, pp. 203-206, 2017. (\url{https://doi.org/10.5195/jmla.2017.88})

\item FORCE11, "Joint Declaration of Data Citation Principles," 2014. (\url{https://doi.org/10.25490/a97f-egyk})

\item DataCite Metadata Working Group, DataCite Metadata Schema Documentation for the Publication and Citation of Research Data, Version 4.5, 2024. (\url{https://doi.org/10.14454/g8e5-6293})

\item A. M. Smith et al., "Software citation principles," PeerJ Computer Science, vol. 2, e86, 2016. (\url{https://doi.org/10.7717/peerj-cs.86})

\item The HDF Group, "Hierarchical Data Format, version 5," 1997-2026. \url{https://www.hdfgroup.org/solutions/hdf5/}

\item R. Rew and G. Davis, "NetCDF: An interface for scientific data access," IEEE Computer Graphics and Applications, vol. 10, no. 4, pp. 76-82, 1990. (\url{https://doi.org/10.1109/38.56302})

\item CF Conventions Committee, "CF metadata conventions," Version 1.11, 2023. \url{https://cfconventions.org/}

\item T. Gebru et al., "Datasheets for datasets," Communications of the ACM, vol. 64, no. 12, pp. 86-92, 2021. (\url{https://doi.org/10.1145/3458723})

\item M. Mitchell et al., "Model cards for model reporting," Proceedings of the Conference on Fairness, Accountability, and Transparency, pp. 220-229, 2019. (\url{https://doi.org/10.1145/3287560.3287596})

\item J. Pineau et al., "Improving reproducibility in machine learning research: A report from the NeurIPS 2019 reproducibility program," Journal of Machine Learning Research, vol. 22, no. 164, pp. 1-20, 2021.

\item S. Kapoor and A. Narayanan, "Leakage and the reproducibility crisis in machine-learning-based science," Patterns, vol. 4, no. 9, 100804, 2023. (\url{https://doi.org/10.1016/j.patter.2023.100804})

\item NED3 Laboratory, "Open-Source Software Packages," Nano Energy and Data-Driven Discovery Laboratory, University of Arkansas. \url{https://ned3.uark.edu/software/}

\item NED3 Laboratory, "Open-Source Datasets," Nano Energy and Data-Driven Discovery Laboratory, University of Arkansas. \url{https://ned3.uark.edu/datasets/}

\item Y. Suh, A. Chandramowlishwaran, and Y. Won, "Recent progress of artificial intelligence for liquid-vapor phase change heat transfer," npj Computational Materials, vol. 10, 65, 2024. (\url{https://doi.org/10.1038/s41524-024-01223-8})

\item G. M. Hobold and A. K. da Silva, "Visualization-based nucleate boiling heat flux quantification using machine learning," International Journal of Heat and Mass Transfer, vol. 134, pp. 511-520, 2019. (\url{https://doi.org/10.1016/j.ijheatmasstransfer.2018.12.170})

\item Y. Suh, R. Bostanabad, and Y. Won, "Deep learning predicts boiling heat transfer," Scientific Reports, vol. 11, 5622, 2021. (\url{https://doi.org/10.1038/s41598-021-85150-4})

\item C.-N. Huang, S. Chang, Y. Suh, I. Mudawar, Y. Won, and C. R. Kharangate, "Machine learning boiling prediction: From autonomous vision of flow visualization data to performance parameter theoretical modeling," International Journal of Multiphase Flow, vol. 179, 104928, 2024. (\url{https://doi.org/10.1016/j.ijmultiphaseflow.2024.104928})

\item A. Bard, Y. Qiu, C. R. Kharangate, and R. H. French, "Consolidated modeling and prediction of heat transfer coefficients for saturated flow boiling in mini/micro-channels using machine learning methods," Applied Thermal Engineering, 118305, 2022. (\url{https://doi.org/10.1016/j.applthermaleng.2022.118305})

\item ywflow, "BubMask: Mask R-CNN for Bubble mask extraction," GitHub repository. \url{https://github.com/ywflow/BubMask}

\item Y. Kim and H. Park, "Deep learning-based automated and universal bubble detection and mask extraction in complex two-phase flows," Scientific Reports, vol. 11, 8940, 2021. (\url{https://doi.org/10.1038/s41598-021-88334-0})

\item C. Dunlap, C. Li, H. Pandey, N. Le, and H. Hu, "BubbleID: A Deep Learning Framework for Bubble Interface Dynamics Analysis," Journal of Applied Physics, vol. 136, 014902, 2024.

\item B. Suriyaprasaad, A. Upadhyay, A. Thakur, and R. Raj, "Explainable boiling acoustics analysis using Grad-CAM and YAMNet for robust pool boiling regime classification," 2025. \url{https://www1.iitp.ac.in/~rraj/datasets.html}

\item H. Pandey, C. Li, and H. Hu, "Multimodal Boiling Dataset with Synchronized Acoustic, Optical, and Thermal Measurements Under Steady-State and Transient Heat Loads," Data in Brief, 55, 110582, 2024.

\item H. Pandey and H. Hu, "Pool Boiling Dataset: Synchronized Acoustic, Optical, and Thermal Measurements for Steady-State and Transient Heat Loads," Harvard Dataverse, 2024. (\url{https://doi.org/10.7910/DVN/6GLGC6})

\item S. Khodakarami, P. Kabirzadeh, C. Wang, T. S. Thukral, and N. Miljkovic, "Optical to infrared mapping of vapor-to-liquid phase change dynamics using generative machine learning," arXiv:2504.17726, 2025. (\url{https://doi.org/10.48550/arXiv.2504.17726})

\item SiaK4, "cGAN-image-to-temperature-mapping," GitHub repository. \url{https://github.com/SiaK4/cGAN-image-to-temperature-mapping}

\item C. Dunlap and contributors, "SeqReg," GitHub repository. \url{https://github.com/cldunlap73/SeqReg}

\item UARK-NED3, "Data-Center-Cooling-Research-Tools," GitHub repository. \url{https://github.com/UARK-NED3/Data-Center-Cooling-Research-Tools}

\item UARK-NED3, "BoilingBench-Multimodal: Seed benchmark for multimodal machine learning with two-phase boiling datasets," GitHub repository, 2026. \url{https://github.com/UARK-NED3/BoilingBench-Multimodal}

\item MultiphaseHUB, "Datahub for Multiphase Transport." \url{https://www.multiphasehub.org/}

\item UARK-NED3, "Bubble-Dynamics-Research-Tools," GitHub repository. \url{https://github.com/UARK-NED3/Bubble-Dynamics-Research-Tools}

\item D. Curl and H. Hu, "Physically Interpretable Surrogate Modeling of Thermal Fields in Electronics Cooling Using Combined Proper Orthogonal Decomposition and Neural Networks," AI Thermal Fluids, 2026.

\item H. Pandey, S. Pierson, C. Li, C. Dunlap, and H. Hu, "Data from: Two-phase immersion cooler for medium-voltage silicon carbide MOSFETs," Dryad pre-publication sharing record, 2025. (\url{https://doi.org/10.5061/dryad.k98sf7mk9})

\item S. Nukiyama, "The maximum and minimum values of the heat Q transmitted from metal to boiling water under atmospheric pressure," Journal of the Japan Society of Mechanical Engineers, vol. 37, no. 206, pp. 367-374, 1934. (\url{https://doi.org/10.1299/jsmemagazine.37.206_367})

\item K. Stephan and M. Abdelsalam, "Heat-transfer correlations for natural convection boiling," International Journal of Heat and Mass Transfer, vol. 23, no. 1, pp. 73-87, 1980. (\url{https://doi.org/10.1016/0017-9310(80)90140-4})

\item S. G. Kandlikar, "A theoretical model to predict pool boiling CHF incorporating effects of contact angle and orientation," Journal of Heat Transfer, vol. 123, no. 6, pp. 1071-1079, 2001. (\url{https://doi.org/10.1115/1.1409265})

\item A. R. Betz, J. Jenkins, C.-J. Kim, and D. Attinger, "Boiling heat transfer on superhydrophilic, superhydrophobic, and superbiphilic surfaces," International Journal of Heat and Mass Transfer, vol. 57, no. 2, pp. 733-741, 2013. (\url{https://doi.org/10.1016/j.ijheatmasstransfer.2012.10.080})

\item K.-H. Chu, R. Enright, and E. N. Wang, "Structured surfaces for enhanced pool boiling heat transfer," Applied Physics Letters, vol. 100, 241603, 2012. (\url{https://doi.org/10.1063/1.4724190})

\item N. S. Dhillon, J. Buongiorno, and K. K. Varanasi, "Critical heat flux maxima during boiling crisis on textured surfaces," Nature Communications, vol. 6, 8247, 2015. (\url{https://doi.org/10.1038/ncomms9247})

\item S. J. Kim, I. C. Bang, J. Buongiorno, and L. W. Hu, "Surface wettability change during pool boiling of nanofluids and its effect on critical heat flux," International Journal of Heat and Mass Transfer, vol. 50, no. 19-20, pp. 4105-4116, 2007. (\url{https://doi.org/10.1016/j.ijheatmasstransfer.2007.02.002})

\item M. M. Rahman, E. Ölçeroğlu, and M. McCarthy, "Scalable nanomanufacturing of virus-templated coatings for enhanced boiling," Advanced Materials Interfaces, vol. 1, no. 2, 1300107, 2014. (\url{https://doi.org/10.1002/admi.201300107})

\item J. C. Chen, "Correlation for boiling heat transfer to saturated fluids in convective flow," Industrial \& Engineering Chemistry Process Design and Development, vol. 5, no. 3, pp. 322-329, 1966. (\url{https://doi.org/10.1021/i260019a023})

\item M. M. Shah, "Chart correlation for saturated boiling heat transfer: Equations and further study," ASHRAE Transactions, vol. 88, pp. 185-196, 1982.

\item K. E. Gungor and R. H. S. Winterton, "A general correlation for flow boiling in tubes and annuli," International Journal of Heat and Mass Transfer, vol. 29, no. 3, pp. 351-358, 1986. (\url{https://doi.org/10.1016/0017-9310(86)90205-X})

\item S. G. Kandlikar, "A general correlation for saturated two-phase flow boiling heat transfer inside horizontal and vertical tubes," Journal of Heat Transfer, vol. 112, no. 1, pp. 219-228, 1990. (\url{https://doi.org/10.1115/1.2910348})

\item J. R. Thome, "Boiling in microchannels: A review of experiment and theory," International Journal of Heat and Fluid Flow, vol. 25, no. 2, pp. 128-139, 2004. (\url{https://doi.org/10.1016/j.ijheatfluidflow.2003.11.005})

\item R. Revellin and J. R. Thome, "A theoretical model for the prediction of the critical heat flux in heated microchannels," International Journal of Heat and Mass Transfer, vol. 51, no. 5-6, pp. 1216-1225, 2008. (\url{https://doi.org/10.1016/j.ijheatmasstransfer.2007.03.039})

\item S. S. Bertsch, E. A. Groll, and S. V. Garimella, "Review and comparative analysis of studies on saturated flow boiling in small channels," Nanoscale and Microscale Thermophysical Engineering, vol. 12, no. 3, pp. 187-227, 2008. (\url{https://doi.org/10.1080/15567260802317357})

\item C. B. Tibirica and G. Ribatski, "Flow boiling heat transfer of R134a and R245fa in a 2.3 mm tube," International Journal of Heat and Mass Transfer, vol. 53, no. 11-12, pp. 2459-2468, 2010. (\url{https://doi.org/10.1016/j.ijheatmasstransfer.2010.01.038})

\item I. Mudawar, "Assessment of high-heat-flux thermal management schemes," IEEE Transactions on Components and Packaging Technologies, vol. 24, no. 2, pp. 122-141, 2001. (\url{https://doi.org/10.1109/6144.926375})

\item L. Breiman, "Random forests," Machine Learning, vol. 45, pp. 5-32, 2001. (\url{https://doi.org/10.1023/A:1010933404324})

\item T. Chen and C. Guestrin, "XGBoost: A scalable tree boosting system," Proceedings of the 22nd ACM SIGKDD International Conference on Knowledge Discovery and Data Mining, pp. 785-794, 2016. (\url{https://doi.org/10.1145/2939672.2939785})

\item G. Ke et al., "LightGBM: A highly efficient gradient boosting decision tree," Advances in Neural Information Processing Systems, vol. 30, 2017.

\item C. E. Rasmussen and C. K. I. Williams, Gaussian Processes for Machine Learning, MIT Press, 2006 (ISBN: 978-0-262-18253-9).

\item V. N. Vapnik, The Nature of Statistical Learning Theory, Springer, 1995 (ISBN: 978-0-387-94559-0).

\item Y. LeCun, Y. Bengio, and G. Hinton, "Deep learning," Nature, vol. 521, pp. 436-444, 2015. (\url{https://doi.org/10.1038/nature14539})

\item I. Goodfellow, Y. Bengio, and A. Courville, Deep Learning, MIT Press, 2016 (ISBN: 978-0-262-03561-3). \url{https://www.deeplearningbook.org/}

\item S. Hochreiter and J. Schmidhuber, "Long short-term memory," Neural Computation, vol. 9, no. 8, pp. 1735-1780, 1997. (\url{https://doi.org/10.1162/neco.1997.9.8.1735})

\item A. Vaswani et al., "Attention is all you need," Advances in Neural Information Processing Systems, vol. 30, 2017.

\item T. Sharma, A. Thakur, and R. Raj, "Exploring acoustic emissions from bubbles in underwater nozzles," 2025. \url{https://www1.iitp.ac.in/~rraj/}

\item A. Krizhevsky, I. Sutskever, and G. E. Hinton, "ImageNet classification with deep convolutional neural networks," Advances in Neural Information Processing Systems, vol. 25, 2012.

\item K. He, X. Zhang, S. Ren, and J. Sun, "Deep residual learning for image recognition," Proceedings of the IEEE Conference on Computer Vision and Pattern Recognition, pp. 770-778, 2016. (\url{https://doi.org/10.1109/CVPR.2016.90})

\item O. Ronneberger, P. Fischer, and T. Brox, "U-Net: Convolutional networks for biomedical image segmentation," Medical Image Computing and Computer-Assisted Intervention, pp. 234-241, 2015. (\url{https://doi.org/10.1007/978-3-319-24574-4_28})

\item K. He, G. Gkioxari, P. Dollár, and R. Girshick, "Mask R-CNN," Proceedings of the IEEE International Conference on Computer Vision, pp. 2961-2969, 2017. (\url{https://doi.org/10.1109/ICCV.2017.322})

\item A. Dosovitskiy et al., "An image is worth 16x16 words: Transformers for image recognition at scale," International Conference on Learning Representations, 2021. \url{https://arxiv.org/abs/2010.11929}

\item J. Redmon, S. Divvala, R. Girshick, and A. Farhadi, "You only look once: Unified, real-time object detection," Proceedings of the IEEE Conference on Computer Vision and Pattern Recognition, pp. 779-788, 2016. (\url{https://doi.org/10.1109/CVPR.2016.91})

\item N. Carion et al., "End-to-end object detection with transformers," European Conference on Computer Vision, pp. 213-229, 2020. (\url{https://doi.org/10.1007/978-3-030-58452-8_13})

\item A. Kirillov et al., "Segment anything," Proceedings of the IEEE/CVF International Conference on Computer Vision, pp. 4015-4026, 2023. (\url{https://doi.org/10.1109/ICCV51070.2023.00371})

\item R. Raj and TFTL, "Datasets \& Code Repositories," Thermal and Fluid Transport Laboratory, Indian Institute of Technology Patna. \url{https://www1.iitp.ac.in/~rraj/datasets.html}

\item G. Gallego et al., "Event-based vision: A survey," IEEE Transactions on Pattern Analysis and Machine Intelligence, vol. 44, no. 1, pp. 154-180, 2022. (\url{https://doi.org/10.1109/TPAMI.2020.3008413})

\item S. L. Brunton, J. L. Proctor, and J. N. Kutz, "Discovering governing equations from data by sparse identification of nonlinear dynamical systems," Proceedings of the National Academy of Sciences, vol. 113, no. 15, pp. 3932-3937, 2016. (\url{https://doi.org/10.1073/pnas.1517384113})

\item M. Raissi, P. Perdikaris, and G. E. Karniadakis, "Physics-informed neural networks: A deep learning framework for solving forward and inverse problems involving nonlinear partial differential equations," Journal of Computational Physics, vol. 378, pp. 686-707, 2019. (\url{https://doi.org/10.1016/j.jcp.2018.10.045})

\item G. E. Karniadakis et al., "Physics-informed machine learning," Nature Reviews Physics, vol. 3, pp. 422-440, 2021. (\url{https://doi.org/10.1038/s42254-021-00314-5})

\item L. Lu, P. Jin, G. Pang, Z. Zhang, and G. E. Karniadakis, "Learning nonlinear operators via DeepONet based on the universal approximation theorem of operators," Nature Machine Intelligence, vol. 3, pp. 218-229, 2021. (\url{https://doi.org/10.1038/s42256-021-00302-5})

\item Z. Li et al., "Fourier neural operator for parametric partial differential equations," International Conference on Learning Representations, 2021. \url{https://arxiv.org/abs/2010.08895}

\item T. Pfaff, M. Fortunato, A. Sanchez-Gonzalez, and P. W. Battaglia, "Learning mesh-based simulation with graph networks," International Conference on Learning Representations, 2021. \url{https://arxiv.org/abs/2010.03409}

\item A. Sanchez-Gonzalez et al., "Learning to simulate complex physics with graph networks," International Conference on Machine Learning, pp. 8459-8468, 2020.

\item D. Kochkov et al., "Machine learning-accelerated computational fluid dynamics," Proceedings of the National Academy of Sciences, vol. 118, no. 21, e2101784118, 2021. (\url{https://doi.org/10.1073/pnas.2101784118})

\item J. Ling, A. Kurzawski, and J. Templeton, "Reynolds averaged turbulence modelling using deep neural networks with embedded invariance," Journal of Fluid Mechanics, vol. 807, pp. 155-166, 2016. (\url{https://doi.org/10.1017/jfm.2016.615})

\item K. Duraisamy, G. Iaccarino, and H. Xiao, "Turbulence modeling in the age of data," Annual Review of Fluid Mechanics, vol. 51, pp. 357-377, 2019. (\url{https://doi.org/10.1146/annurev-fluid-010518-040547})

\item S. L. Brunton, B. R. Noack, and P. Koumoutsakos, "Machine learning for fluid mechanics," Annual Review of Fluid Mechanics, vol. 52, pp. 477-508, 2020. (\url{https://doi.org/10.1146/annurev-fluid-010719-060214})

\item J. Willard, X. Jia, S. Xu, M. Steinbach, and V. Kumar, "Integrating scientific knowledge with machine learning for engineering and environmental systems," ACM Computing Surveys, vol. 55, no. 4, 66, 2022. (\url{https://doi.org/10.1145/3514228})

\item R. Vinuesa and S. L. Brunton, "Enhancing computational fluid dynamics with machine learning," Nature Computational Science, vol. 2, pp. 358-366, 2022. (\url{https://doi.org/10.1038/s43588-022-00264-7})

\item F. Al-Hindawi, T. Soori, H. Hu, M. M. R. Siddiquee, H. Yoon, T. Wu, and Y. Sun, "A framework for generalizing critical heat flux detection models using unsupervised image-to-image translation," Expert Systems with Applications, vol. 227, 120265, 2023. (\url{https://doi.org/10.1016/j.eswa.2023.120265})

\item F. Al-Hindawi, M. M. R. Siddiquee, A. Patharkar, J. Huang, T. Wu, and H. Hu, "SequenceSync-GAN: sequence-aware domain translation for generalizable boiling crisis detection," Signal, Image and Video Processing, vol. 20, no. 5, p. 257, 2026. (\url{https://doi.org/10.1007/s11760-026-05338-x})

\item F. Al-Hindawi, "Addressing the Domain Shift in Computer Vision for Boiling Crisis Detection," Ph.D. dissertation, Arizona State University, Tempe, AZ, USA, 2024. \url{https://www.proquest.com/dissertations-theses/addressing-domain-shift-computer-vision-boiling/docview/3122900683/se-2}

\item S. M. Rassoulinejad-Mousavi, F. Al-Hindawi, T. Soori, A. Rokoni, H. Yoon, H. Hu, T. Wu, and Y. Sun, "Deep learning strategies for critical heat flux detection in pool boiling," Applied Thermal Engineering, vol. 190, 116849, 2021. (\url{https://doi.org/10.1016/j.applthermaleng.2021.116849})

\item F. Al-Hindawi, M. M. R. Siddiquee, T. Wu, H. Hu, and Y. Sun, "Domain-knowledge Inspired Pseudo Supervision (DIPS) for Unsupervised Image-to-Image Translation Models to Support Cross-Domain Classification," Engineering Applications of Artificial Intelligence, vol. 127, 107255, 2024. (\url{https://doi.org/10.1016/j.engappai.2023.107255})

\item J. V. Beck, B. Blackwell, and C. R. St. Clair, Jr., \textit{Inverse Heat Conduction: Ill-Posed Problems}, Wiley-Interscience, 1985.

\item A. N. Tikhonov and V. Y. Arsenin, \textit{Solutions of Ill-Posed Problems}, Winston, Washington, DC, 1977.

\item J. Wang and N. Zabaras, ``A Bayesian inference approach to the inverse heat conduction problem,'' \textit{International Journal of Heat and Mass Transfer}, vol. 47, no. 17--18, pp. 3927--3941, 2004. (\url{https://doi.org/10.1016/j.ijheatmasstransfer.2004.02.028})

\item Q. He, D. Barajas-Solano, G. Tartakovsky, and A. M. Tartakovsky, ``Physics-informed neural networks for multiphysics data assimilation with application to subsurface transport,'' \textit{Advances in Water Resources}, vol. 141, 103610, 2020. (\url{https://doi.org/10.1016/j.advwatres.2020.103610})

\item R. Olabiyi, H. Pandey, H. Hu, and A. Iquebal, "A Bayesian spatiotemporal modeling approach to the inverse heat conduction problem," ASME Journal of Heat and Mass Transfer, vol. 146, no. 9, 091403, 2024. (\url{https://doi.org/10.1115/1.4065451})

\item U. Fasel, J. N. Kutz, B. W. Brunton, and S. L. Brunton, ``Ensemble-SINDy: Robust sparse model discovery in the low-data, high-noise limit, with active learning and control,'' \textit{Proceedings of the Royal Society A}, vol. 478, no. 2260, 20210904, 2022. (\url{https://doi.org/10.1098/rspa.2021.0904})

\item D. A. Messenger and D. M. Bortz, ``Weak SINDy for partial differential equations,'' \textit{Journal of Computational Physics}, vol. 443, 110525, 2021. (\url{https://doi.org/10.1016/j.jcp.2021.110525})

\item S. M. Hirsh, D. A. Barajas-Solano, and J. N. Kutz, ``Sparsifying priors for Bayesian uncertainty quantification in model discovery,'' \textit{Royal Society Open Science}, vol. 9, no. 2, 211823, 2022. (\url{https://doi.org/10.1098/rsos.211823})

\item L. S.-T. Fung, U. Fasel, and J. N. Kutz, ``Rapid Bayesian identification of sparse nonlinear dynamics from scarce and noisy data,'' arXiv:2402.15357, 2025.

\item R. Olabiyi, H. Hu, and A. Iquebal, "Discovering governing equations in the presence of uncertainty," arXiv:2507.09740, 2025. (\url{https://doi.org/10.48550/arXiv.2507.09740})

\item E. Carlstein, K.-A. Do, P. Hall, T. Hesterberg, and H. R. K{\"u}nsch, ``Matched-block bootstrap for dependent data,'' \textit{Bernoulli}, vol. 4, no. 3, pp. 305--328, 1998. (\url{https://doi.org/10.2307/3318714})

\item D. Qu{\'e}r{\'e}, ``Inertial capillarity,'' \textit{Europhysics Letters}, vol. 39, no. 5, pp. 533--538, 1997. (\url{https://doi.org/10.1209/epl/i1997-00389-2})

\item Y. Qiu, D. Garg, L. Zhou, C. R. Kharangate, S.-M. Kim, and I. Mudawar, "An artificial neural network model to predict mini/micro-channels saturated flow boiling heat transfer coefficient based on universal consolidated data," International Journal of Heat and Mass Transfer, vol. 149, 119211, 2020. (\url{https://doi.org/10.1016/j.ijheatmasstransfer.2019.119211})

\item L. Zhou, D. Garg, Y. Qiu, S.-M. Kim, I. Mudawar, and C. R. Kharangate, "Machine learning algorithms to predict flow condensation heat transfer coefficient in mini/micro-channel utilizing universal data," International Journal of Heat and Mass Transfer, vol. 162, 120351, 2020. (\url{https://doi.org/10.1016/j.ijheatmasstransfer.2020.120351})

\item M. T. Hughes, B. M. Fronk, and S. Garimella, "Universal condensation heat transfer and pressure drop model and the role of machine learning techniques to improve predictive capabilities," International Journal of Heat and Mass Transfer, vol. 179, 121712, 2021. (\url{https://doi.org/10.1016/j.ijheatmasstransfer.2021.121712})

\item M. T. Hughes, S. M. Chen, and S. Garimella, "Machine-learning-based heat transfer and pressure drop model for internal flow condensation of binary mixtures," International Journal of Heat and Mass Transfer, vol. 194, 123109, 2022. (\url{https://doi.org/10.1016/j.ijheatmasstransfer.2022.123109})

\item M. Bongarala, J. A. Weibel, and S. V. Garimella, "A method to partition boiling heat transfer mechanisms using synchronous through-substrate high-speed visual and infrared measurements," International Journal of Heat and Mass Transfer, vol. 226, 125516, 2024. (\url{https://doi.org/10.1016/j.ijheatmasstransfer.2024.125516})

\item T. H. N. Vu, N. T. Vu, Y. Huang, and J. A. Weibel, "High-speed visualization and infrared thermography of pool boiling on surfaces having differing dynamic wettability," International Journal of Heat and Mass Transfer, vol. 221, 125105, 2024. (\url{https://doi.org/10.1016/j.ijheatmasstransfer.2023.125105})

\item X. Ma and W. Pan, "Data-driven nonintrusive reduced order modeling for dynamical systems with moving boundaries using Gaussian process regression," Computer Methods in Applied Mechanics and Engineering, vol. 373, 113495, 2021. (\url{https://doi.org/10.1016/j.cma.2020.113495})

\item L. Zhu, Z. J. Ooi, T. Zhang, C. S. Brooks, and L. Pan, "Identification of flow regimes in boiling flow with clustering algorithms: An interpretable machine-learning perspective," Applied Thermal Engineering, vol. 228, 120493, 2023. (\url{https://doi.org/10.1016/j.applthermaleng.2023.120493})

\item M. K. Das and N. Kishor, "Adaptive fuzzy model identification to predict the heat transfer coefficient in pool boiling of distilled water," Expert Systems with Applications, vol. 36, no. 2, pp. 1142-1154, 2009. (\url{https://doi.org/10.1016/j.eswa.2007.10.044})

\item N. Vaziri, A. Hojabri, A. Erfani, M. Monsefi, and B. Nilforooshan, "Critical heat flux prediction by using radial basis function and multilayer perceptron neural networks: A comparison study," Nuclear Engineering and Design, vol. 237, no. 4, pp. 377-385, 2007. (\url{https://doi.org/10.1016/j.nucengdes.2006.05.005})

\item X. Zhao, K. Shirvan, R. K. Salko, and F. Guo, "On the prediction of critical heat flux using a physics-informed machine learning-aided framework," Applied Thermal Engineering, vol. 164, 114540, 2020. (\url{https://doi.org/10.1016/j.applthermaleng.2019.114540})

\item W. Zhou, S. Miwa, H. Wang, and K. Okamoto, "Assessment of the state-of-the-art AI methods for critical heat flux prediction," International Communications in Heat and Mass Transfer, vol. 158, 107844, 2024. (\url{https://doi.org/10.1016/j.icheatmasstransfer.2024.107844})

\item B. Swartz, L. Wu, Q. Zhou, and Q. Hao, "Machine learning predictions of critical heat fluxes for pillar-modified surfaces," International Journal of Heat and Mass Transfer, vol. 180, 121744, 2021. (\url{https://doi.org/10.1016/j.ijheatmasstransfer.2021.121744})

\item D. Lu, Y. Suh, and Y. Won, "Rapid identification of boiling crisis with event-based visual streaming analysis," Applied Thermal Engineering, vol. 239, 122004, 2024. (\url{https://doi.org/10.1016/j.applthermaleng.2023.122004})

\item S. Chang, Y. Suh, C. Shingote, C. N. Huang, I. Mudawar, C. Kharangate, and Y. Won, "BubbleMask: Autonomous visualization of digital flow bubbles for predicting critical heat flux," International Journal of Heat and Mass Transfer, vol. 217, 124656, 2023. (\url{https://doi.org/10.1016/j.ijheatmasstransfer.2023.124656})

\item J. H. Seong, M. Ravichandran, G. Su, B. Phillips, and M. Bucci, "Automated bubble analysis of high-speed subcooled flow boiling images using U-net transfer learning and global optical flow," International Journal of Multiphase Flow, vol. 159, 104336, 2023. (\url{https://doi.org/10.1016/j.ijmultiphaseflow.2022.104336})

\item N. Ali, B. Viggiano, M. Tutkun, and R. B. Cal, "Cluster-based reduced-order descriptions of two phase flows," Chemical Engineering Science, vol. 222, 115660, 2020. (\url{https://doi.org/10.1016/j.ces.2020.115660})

\item Z. J. Ooi, L. Zhu, J. L. Bottini, and C. S. Brooks, "Identification of flow regimes in boiling flows in a vertical annulus channel with machine learning techniques," International Journal of Heat and Mass Transfer, vol. 185, 122439, 2022. (\url{https://doi.org/10.1016/j.ijheatmasstransfer.2021.122439})

\item G. M. Hobold and A. K. da Silva, "Automatic detection of the onset of film boiling using convolutional neural networks and Bayesian statistics," International Journal of Heat and Mass Transfer, vol. 134, pp. 262-270, 2019. (\url{https://doi.org/10.1016/j.ijheatmasstransfer.2018.12.070})

\item Y. Liu, N. Dinh, Y. Sato, and B. Niceno, "Data-driven modeling for boiling heat transfer: Using deep neural networks and high-fidelity simulation results," Applied Thermal Engineering, vol. 144, pp. 305-320, 2018. (\url{https://doi.org/10.1016/j.applthermaleng.2018.08.041})

\item S. Khodakarami, K. Fazle Rabbi, Y. Suh, Y. Won, and N. Miljkovic, "Machine learning enabled condensation heat transfer measurement," International Journal of Heat and Mass Transfer, vol. 194, 123016, 2022. (\url{https://doi.org/10.1016/j.ijheatmasstransfer.2022.123016})

\item O. Russakovsky et al., "ImageNet Large Scale Visual Recognition Challenge," International Journal of Computer Vision, vol. 115, pp. 211-252, 2015. (\url{https://doi.org/10.1007/s11263-015-0816-y})

\item T.-Y. Lin et al., "Microsoft COCO: Common Objects in Context," European Conference on Computer Vision, pp. 740-755, 2014. (\url{https://doi.org/10.1007/978-3-319-10602-1_48})

\item L. Bossard, M. Guillaumin, and L. Van Gool, "Food-101: Mining Discriminative Components with Random Forests," European Conference on Computer Vision, pp. 446-461, 2014. (\url{https://doi.org/10.1007/978-3-319-10599-4_29})

\item W. Wu, T. Chen, H. Liu, and L. Cao, "A large-scale benchmark for food image segmentation," Proceedings of the 29th ACM International Conference on Multimedia, pp. 506-515, 2021. (\url{https://doi.org/10.1145/3474085.3475201})

\item A. Mandlekar et al., "What matters in learning from offline human demonstrations for robot manipulation," Conference on Robot Learning, 2021. \url{https://arxiv.org/abs/2108.03298}

\item M. Kang and B. Kwon, "Deep learning of forced convection heat transfer," Journal of Heat Transfer, vol. 144, no. 2, 021801, 2022. (\url{https://doi.org/10.1115/1.4052893})

\item M. T. Hughes, G. Kini, and S. Garimella, "Status, challenges, and potential for machine learning in understanding and applying heat transfer phenomena," Journal of Heat Transfer, vol. 143, no. 12, 120802, 2021. (\url{https://doi.org/10.1115/1.4052510})

\item B. Kwon, F. Ejaz, and L. K. Hwang, "Machine learning for heat transfer correlations," International Communications in Heat and Mass Transfer, vol. 116, 104694, 2020. (\url{https://doi.org/10.1016/j.icheatmasstransfer.2020.104694})

\item L. Lin, L. Gao, M. A. Kedzierski, and Y. Hwang, "A general model for flow boiling heat transfer in microfin tubes based on a new neural network architecture," Energy and AI, vol. 8, 100151, 2022. (\url{https://doi.org/10.1016/j.egyai.2022.100151})

\item Y. Kuang, F. Han, L. Sun, R. Zhuan, and W. Wang, "Saturated hydrogen nucleate flow boiling heat transfer coefficients study based on artificial neural network," International Journal of Heat and Mass Transfer, vol. 175, 121406, 2021. (\url{https://doi.org/10.1016/j.ijheatmasstransfer.2021.121406})

\item R. Wajman, "Computer methods for non-invasive measurement and control of two-phase flows: A review study," Information Technology and Control, vol. 48, no. 3, pp. 464-486, 2019. (\url{https://doi.org/10.5755/j01.itc.48.3.22189})

\item H. B. Arteaga-Arteaga et al., "Machine learning applications to predict two-phase flow patterns," PeerJ Computer Science, vol. 7, e798, 2021. (\url{https://doi.org/10.7717/peerj-cs.798})

\item S. Passoni, R. Mereu, and M. Bucci, "Integrating machine learning and image processing for void fraction estimation in two-phase flow through corrugated channels," International Journal of Multiphase Flow, vol. 177, 104871, 2024. (\url{https://doi.org/10.1016/j.ijmultiphaseflow.2024.104871})

\item N. V. Upot, K. Fazle Rabbi, S. Khodakarami, J. Y. Ho, J. Kohler Mendizabal, and N. Miljkovic, "Advances in micro and nanoengineered surfaces for enhancing boiling and condensation heat transfer: A review," Nanoscale Advances, vol. 5, no. 5, pp. 1232-1270, 2023. (\url{https://doi.org/10.1039/d2na00669c})

\item K. H. Chu, Y. S. Joung, R. Enright, C. R. Buie, and E. N. Wang, "Hierarchically structured surfaces for boiling critical heat flux enhancement," Applied Physics Letters, vol. 102, no. 15, 151602, 2013. (\url{https://doi.org/10.1063/1.4801811})

\item J. Li, W. Fu, B. Zhang, G. Zhu, and N. Miljkovic, "Ultrascalable three-tier hierarchical nanoengineered surfaces for optimized boiling," ACS Nano, vol. 13, no. 12, pp. 14080-14093, 2019. (\url{https://doi.org/10.1021/acsnano.9b06501})

\item M. M. Rahman, E. Olceroglu, and M. McCarthy, "Role of wickability on the critical heat flux of structured superhydrophilic surfaces," Langmuir, vol. 30, no. 37, pp. 11225-11234, 2014. (\url{https://doi.org/10.1021/la5030923})

\item S. H. Kim, G. C. Lee, J. Y. Kang, K. Moriyama, M. H. Kim, and H. S. Park, "Boiling heat transfer and critical heat flux evaluation of the pool boiling on micro structured surface," International Journal of Heat and Mass Transfer, vol. 91, pp. 1140-1147, 2015. (\url{https://doi.org/10.1016/j.ijheatmasstransfer.2015.07.120})

\item Z. G. Xu, Z. G. Qu, C. Y. Zhao, and W. Q. Tao, "Pool boiling heat transfer on open-celled metallic foam sintered surface under saturation condition," International Journal of Heat and Mass Transfer, vol. 54, no. 17-18, pp. 3856-3867, 2011. (\url{https://doi.org/10.1016/j.ijheatmasstransfer.2011.04.043})

\item M. Calati, G. Righetti, L. Doretti, C. Zilio, G. A. Longo, K. Hooman, and S. Mancin, "Water pool boiling in metal foams: From experimental results to a generalized model based on artificial neural network," International Journal of Heat and Mass Transfer, vol. 176, 121451, 2021. (\url{https://doi.org/10.1016/j.ijheatmasstransfer.2021.121451})

\item D. E. Kim, D. I. Yu, S. C. Park, H. J. Kwak, and H. S. Ahn, "Critical heat flux triggering mechanism on micro-structured surfaces: Coalesced bubble departure frequency and liquid furnishing capability," International Journal of Heat and Mass Transfer, vol. 91, pp. 1237-1247, 2015. (\url{https://doi.org/10.1016/j.ijheatmasstransfer.2015.08.065})

\item S. Jun, J. Kim, D. Son, H. Y. Kim, and S. M. You, "Enhancement of pool boiling heat transfer in water using sintered copper microporous coatings," Nuclear Engineering and Technology, vol. 48, no. 4, pp. 932-940, 2016. (\url{https://doi.org/10.1016/j.net.2016.02.018})

\item J. Li, G. Zhu, D. Kang, W. Fu, Y. Zhao, and N. Miljkovic, "Endoscopic visualization of contact line dynamics during pool boiling on capillary-activated copper microchannels," Advanced Functional Materials, vol. 31, no. 4, 2006249, 2021. (\url{https://doi.org/10.1002/adfm.202006249})

\item H. X. Chen, Y. Sun, L. H. Li, and X. D. Wang, "Bubble dynamics and heat transfer performance on micro-pillars structured surfaces with various pillars heights," International Journal of Heat and Mass Transfer, vol. 163, 120502, 2020. (\url{https://doi.org/10.1016/j.ijheatmasstransfer.2020.120502})

\item Q. N. Pham, S. Zhang, S. Hao, K. Montazeri, C. H. Lin, J. Lee, A. Mohraz, and Y. Won, "Boiling heat transfer with a well-ordered microporous architecture," ACS Applied Materials \& Interfaces, vol. 12, no. 16, pp. 19174-19183, 2020. (\url{https://doi.org/10.1021/acsami.0c01113})

\item D. I. Yu, H. J. Kwak, H. Noh, H. S. Park, K. Fezzaa, and M. H. Kim, "Synchrotron x-ray imaging visualization study of capillary-induced flow and critical heat flux on surfaces with engineered micropillars," Science Advances, vol. 4, no. 2, e1701571, 2018. (\url{https://doi.org/10.1126/sciadv.1701571})

\item L. Zhang, J. H. Seong, and M. Bucci, "Percolative scale-free behavior in the boiling crisis," Physical Review Letters, vol. 122, no. 13, 134501, 2019. (\url{https://doi.org/10.1103/PhysRevLett.122.134501})

\item M. C. Lu, R. Chen, V. Srinivasan, V. P. Carey, and A. Majumdar, "Critical heat flux of pool boiling on Si nanowire array-coated surfaces," International Journal of Heat and Mass Transfer, vol. 54, no. 25-26, pp. 5359-5367, 2011. (\url{https://doi.org/10.1016/j.ijheatmasstransfer.2011.08.007})

\item D. Lee, N. Lee, D. I. Shim, B. S. Kim, and H. H. Cho, "Enhancing thermal stability and uniformity in boiling heat transfer using micro-nano hybrid surfaces (MNHS)," Applied Thermal Engineering, vol. 130, pp. 710-721, 2018. (\url{https://doi.org/10.1016/j.applthermaleng.2017.10.144})

\item G. Pi, D. Deng, L. Chen, X. Xu, and C. Zhao, "Pool boiling performance of 3D-printed reentrant microchannels structures," International Journal of Heat and Mass Transfer, vol. 156, 119920, 2020. (\url{https://doi.org/10.1016/j.ijheatmasstransfer.2020.119920})

\item S. Xie, M. Jiang, H. Kong, Q. Tong, and J. Zhao, "An experimental investigation on the pool boiling of multi-orientated hierarchical structured surfaces," International Journal of Heat and Mass Transfer, vol. 164, 120595, 2021. (\url{https://doi.org/10.1016/j.ijheatmasstransfer.2020.120595})

\item A. R. Betz, J. Xu, H. Qiu, and D. Attinger, "Do surfaces with mixed hydrophilic and hydrophobic areas enhance pool boiling?," Applied Physics Letters, vol. 97, no. 14, 141909, 2010. (\url{https://doi.org/10.1063/1.3485057})

\item R. Chen, M. C. Lu, V. Srinivasan, Z. Wang, H. H. Cho, and A. Majumdar, "Nanowires for enhanced boiling heat transfer," Nano Letters, vol. 9, no. 2, pp. 548-553, 2009. (\url{https://doi.org/10.1021/nl8026857})

\item A. Elkholy and R. Kempers, "Enhancement of pool boiling heat transfer using 3D-printed polymer fixtures," Experimental Thermal and Fluid Science, vol. 114, 110056, 2020. (\url{https://doi.org/10.1016/j.expthermflusci.2020.110056})

\item D. C. Mo, S. Yang, J. L. Luo, Y. Q. Wang, and S. S. Lyu, "Enhanced pool boiling performance of a porous honeycomb copper surface with radial diameter gradient," International Journal of Heat and Mass Transfer, vol. 157, 119867, 2020. (\url{https://doi.org/10.1016/j.ijheatmasstransfer.2020.119867})

\item L. Mao, W. Zhou, X. Hu, Y. He, G. Zhang, L. Zhang, and R. Fu, "Pool boiling performance and bubble dynamics on graphene oxide nanocoating surface," International Journal of Thermal Sciences, vol. 147, 106154, 2020. (\url{https://doi.org/10.1016/j.ijthermalsci.2019.106154})
\end{enumerate}
\end{document}